\documentclass[10pt,twocolumn,letterpaper]{article}

\usepackage{wacv}
\usepackage{times}
\usepackage{epsfig}
\usepackage{graphicx}
\usepackage{amsmath}
\usepackage{amssymb}
\usepackage{booktabs}

\usepackage{array, multirow}
\usepackage{mathtools}
\usepackage{enumitem}
\usepackage{float}
\usepackage[normalem]{ulem}
\useunder{\uline}{\ul}{}
\graphicspath{ {./pics/} }
\usepackage{bbm}

\usepackage[T1]{fontenc} % correctly show < and >
\usepackage[dvipsnames]{xcolor}

\usepackage{pifont}
\newcommand{\cmark}{\ding{51}}%

\usepackage{tikz}
\usetikzlibrary{calc}
\usetikzlibrary{shapes}

\definecolor{fig1-orange}{RGB}{237, 125, 49}
\definecolor{fig1-blue}{RGB}{68, 114, 196}
\definecolor{fig1-purple}{RGB}{112, 48, 160}
\definecolor{fig1-pink}{RGB}{204, 0, 153}

\makeatletter
\newdimen\@myBoxHeight%
\newdimen\@myBoxDepth%
\newdimen\@myBoxWidth%
\newdimen\@myBoxSize%
\newcommand{\SquareBox}[2][]{%
    \settoheight{\@myBoxHeight}{#2}% Record height of box
    \settodepth{\@myBoxDepth}{#2}% Record depth of box
    \settowidth{\@myBoxWidth}{#2}% Record width of box
    \pgfmathsetlength{\@myBoxSize}{max(\@myBoxWidth,(\@myBoxHeight+\@myBoxDepth))}%
    \tikz \node [shape=rectangle, shape aspect=1,draw=red,inner sep=2\pgflinewidth, minimum size=\@myBoxSize,#1] {#2};%
}%
\makeatother

\usepackage[accsupp]{axessibility}

\def\wacvPaperID{1046} % Enter the WACV Paper ID here

\wacvalgorithmstrack   % Uncomment this line if you are submitting to the Algorithms Track.
\wacvfinalcopy % *** Uncomment this line for the final submission

\ifwacvfinal
\usepackage[breaklinks=true,bookmarks=false]{hyperref}
\else
\usepackage[pagebackref=true,breaklinks=true,colorlinks,bookmarks=false]{hyperref}
\fi

\begin{document}

%%%%%%%%% TITLE
\title{Detection Recovery in Online Multi-Object Tracking with Sparse Graph Tracker}

\renewcommand{\thefootnote}{\fnsymbol{footnote}}
\author{
Jeongseok Hyun\textsuperscript{1}\footnotemark[1] \qquad Myunggu Kang\textsuperscript{2} \qquad Dongyoon Wee\textsuperscript{2} \qquad Dit-Yan Yeung\textsuperscript{1}\\
\textsuperscript{1}The Hong Kong University of Science and Technology \qquad \textsuperscript{2}Clova AI, NAVER Corp.\\
}

\maketitle
\thispagestyle{empty} % uncomment for camera-ready to suppress page number
\footnotetext[1]{The work was partially done during an intern at Clova AI.}

\begin{abstract}
In existing joint detection and tracking methods, pairwise relational features are used to match previous tracklets to current detections.
However, the features may not be discriminative enough for a tracker to identify a target from a large number of detections.
Selecting only high-scored detections for tracking may lead to missed detections whose confidence score is low.
Consequently, in the online setting, this results in disconnections of tracklets which cannot be recovered.
In this regard, we present Sparse Graph Tracker (SGT), a novel online graph tracker using higher-order relational features which are more discriminative by aggregating the features of neighboring detections and their relations. 
SGT converts video data into a graph where detections, their connections, and the relational features of two connected nodes are represented by nodes, edges, and edge features, respectively.
The strong edge features allow SGT to track targets with tracking candidates selected by top-$K$ scored detections with large $K$.
As a result, even low-scored detections can be tracked, and the missed detections are also recovered.
The robustness of $K$ value is shown through the extensive experiments.
In the MOT16/17/20 and HiEve Challenge, SGT outperforms the state-of-the-art trackers with real-time inference speed. 
Especially, a large improvement in MOTA is shown in the MOT20 and HiEve Challenge. 
Code is available at \href{https://github.com/HYUNJS/SGT}{\textcolor{cyan}{https://github.com/HYUNJS/SGT}}.
\end{abstract}

\section{Introduction}
\label{sec:intro}

\begin{figure}[t!]
    \centering
    \includegraphics[width=\linewidth]{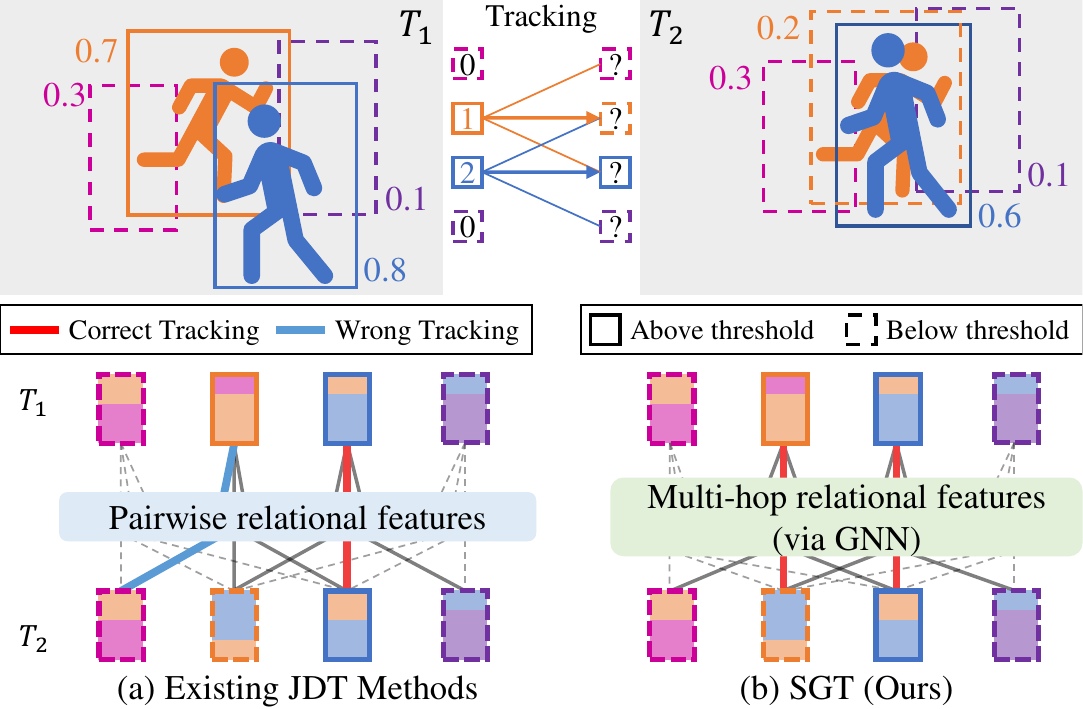}
    \caption{
    \textbf{Motivation of detection recovery by tracking.} 
    Tracking of \SquareBox[draw=fig1-orange, thick]{\textcolor{fig1-orange}{\scriptsize 1}} cannot be performed if the high-scored detection (\SquareBox[draw=fig1-blue, thick]{\scriptsize ?}) is selected as the tracking candidate.
    Meanwhile, adding low-scored detections (\SquareBox[draw=fig1-orange, thick, dashed]{\scriptsize ?}, \SquareBox[draw=fig1-purple, thick, dashed]{\scriptsize ?}, \SquareBox[draw=fig1-pink, thick, dashed]{\scriptsize ?}) to the candidate pool results in wrong matching of \SquareBox[draw=fig1-orange, thick]{\textcolor{fig1-orange}{\scriptsize 1}} with \SquareBox[draw=fig1-pink, thick, dashed]{\scriptsize ?} in~\textbf{(a)} since pairwise relations are ambiguous to be used for discriminating \SquareBox[draw=fig1-orange, thick]{\textcolor{fig1-orange}{\scriptsize 1}} among them.
    In contrast, our SGT~\textbf{(b)} exploits multi-hop relations updated via a GNN to contain visual features of neighboring detections and their relations. Despite many tracking candidates selected by top-$K$, SGT successfully tracks \SquareBox[draw=fig1-orange, thick]{\textcolor{fig1-orange}{\scriptsize 1}}, and its missed detection is consequently recovered.
    }
    \label{fig:sgt_motivation}
\end{figure}

In the online setting, missed detection problem is far more critical than in the offline setting; tracklets are disconnected once the corresponding detections are missed, while tracklet interpolation is infeasible to fill the past missed detections. 
As illustrated in Figure~\ref{fig:sgt_motivation}, occlusion leads to low-confident detections, and if they are included in the association step, the complexity of tracking increases with too many spurious detections. 
Pairwise relational features (\eg,~position or visual similarity) may not be discriminative enough to distinguish targets in such case and result in wrong matching.
Thus, existing works~\cite{fairmot, gtr, cstrack, jde}, which use the pairwise relations for tracking, exploit only high-scored detections as tracking candidates.

A graph is an effective way to represent relations between objects in a video, and a graph neural network (GNN) is effective in modeling the relationship. 
Bearing this in mind, we model the spatio-temporal relationship in video data using a GNN to extract higher-order relational features (\ie,~multi-hop relational features) which consider the relations between neighboring objects or background patches. 
These features are powerful and can perform association correctly even if a large number of detections (\eg,~300) is selected as tracking candidates. 
Thus, we propose Sparse Graph Tracker (SGT), a novel online graph tracker that adopts joint detection and tracking (JDT) framework~\cite{jde} where object detector and tracker share a backbone network to achieve fast inference speed. 

In the association step, existing online JDT methods utilize pairwise relational features between two detections, such as similarity of appearance features~\cite{retinatrack, jde, fairmot, gsdt}, center point distance~\cite{centertrack}, and Intersection over Union (IoU) score~\cite{transtrack}. 
Although \cite{deepsort, jde, fairmot, gsdt} fuse appearance information and motion information by weighted sum, these features reflect only the relations between two objects and are not discriminative for accurate matching in a crowded scene. 
Motion predictor (\eg,~Kalman filter~\cite{kalman_filter}) is commonly employed to improve tracking performance. 
On the contrary, we utilize higher-order relational features by aggregating the features of neighboring nodes and edges through iterations of GNNs. 
Even without a motion predictor, higher-order relational features are still powerful to correctly match the previous tracklets with current frame's ($I_{t2}$) top-$K$ scored detections which contain a large number of spurious detections due to a large $K$ value.

%% our contribution
The main contributions of this work are as follows: 
\begin{enumerate}[noitemsep,topsep=0pt,leftmargin=2.5em]
    \item We propose a novel online graph-based tracker that is jointly trained with object detector and performs long-term association without any motion model. Our SGT shows superior performance on the MOT16/17/20 and HiEve benchmarks with real-time inference speed.
    \item We propose training and inference techniques for SGT effectively achieving detection recovery by tracking. Their effectiveness is demonstrated through extensive ablation experiments and a large improvement on MOT20 where severe crowdedness results in low confident detections for the occluded objects.
\end{enumerate}

\section{Related Works}
\label{sec:related_work}

\subsection{JDT Methods}

Recently, many JDT methods are proposed due to fast inference speed and single-stage training based on the shared backbone. They extend object detectors to MOT models with the extra tracking branch which is jointly trained with the detector. They fall into two categories: (1)~re-identification (reID) branch outputs discriminative appearance features for tracking, and (2)~motion prediction branch outputs the updated location of tracklets for tracking.

\noindent \textbf{JDT by reID.} 
RetinaTrack~\cite{retinatrack}, JDE~\cite{jde}, FairMOT~\cite{fairmot} append reID branch to RetinaNet~\cite{focal_loss}, YOLOv3~\cite{yolov3}, and CenterNet~\cite{centernet}, respectively. Liang~\etal~\cite{cstrack} points out that the objectives of detection and ReID are conflicting, and proposes a cross-correlation network that learns task-specific features. On the other hand, GSDT~\cite{gsdt} and CorrTracker~\cite{corrtracker} enhance the current features by spatio-temporal relational modeling that exploits previous frames. CorrTracker~\cite{corrtracker}, the current state-of-ther-art model, fuses the correlation in both spatial and temporal dimensions to the image features at multiple pyramid levels. All these methods associate the tracklets and detections using the similarity of reID features. Also, Kalman filter~\cite{kalman_filter} is commonly employed and motion information is fused to the similarity.

\noindent \textbf{JDT by motion prediction.} 
D\&T~\cite{d2t2d} and CenterTrack~\cite{centertrack} append the learnable motion predictor into R-FCN~\cite{r-fcn} and CenterNet~\cite{centernet}, respectively. CenterTrack associates the tracklets and detections using the center point distance of the detections and the tracklets updated by predicted motion. TraDeS~\cite{trades} predicts the center offset of objects between two consecutive frames based on a cost volume which is computed by a similarity of reID features of the frames. TransTrack~\cite{transtrack} is a transformer-based tracker that propagates the previous frame's tracklets to the current frame and matches with the current detections by IoU score.

\noindent \textbf{Comparison.} Our SGT extends CenterNet~\cite{centernet} with a graph tracker. Compared with others using pairwise relational features (\eg,~IoU, cosine similarity or fusion of them), SGT exploits edge features updated through GNN which are higher-order relational features, and solves association as edge classification as shown in Figure~\ref{fig:sgt_motivation}.

\subsection{Graph-based Multi-object Tracking}

A graph is an effective way to represent relational information, and GNN can learn higher-order relational information through a message passing process that propagates node or edge features to the connected nodes or edges and aggregates neighboring features. STRN~\cite{strn} is an online MOT method with a spatio-temporal relation network that consists of a spatial relation module and a temporal relation module. The features from these two modules are fused to predict the affinity score for association. MPNTrack~\cite{tmpnn} adopts a message passing network~\cite{mpnn} with time-aware node update module that aggregates past and future features separately and solves MOT problem as edge classification. LPCMOT~\cite{lpcmot} generates and scores tracklet proposals based on a set of frames and detections with Graph Convolution Network (GCN)~\cite{gcn}. GSDT~\cite{gsdt} is the first work that applies a GNN in an online JDT method, but its use of GNN is limited to enhancing the current feature map and tracking is still performed using pairwise relational features. In contrast, SGT is the first JDT method using higher-order relational features for tracking.

\begin{figure*}[t] %% place it here to put figure into page 3
    \centering
    \includegraphics[width=1.0\textwidth]{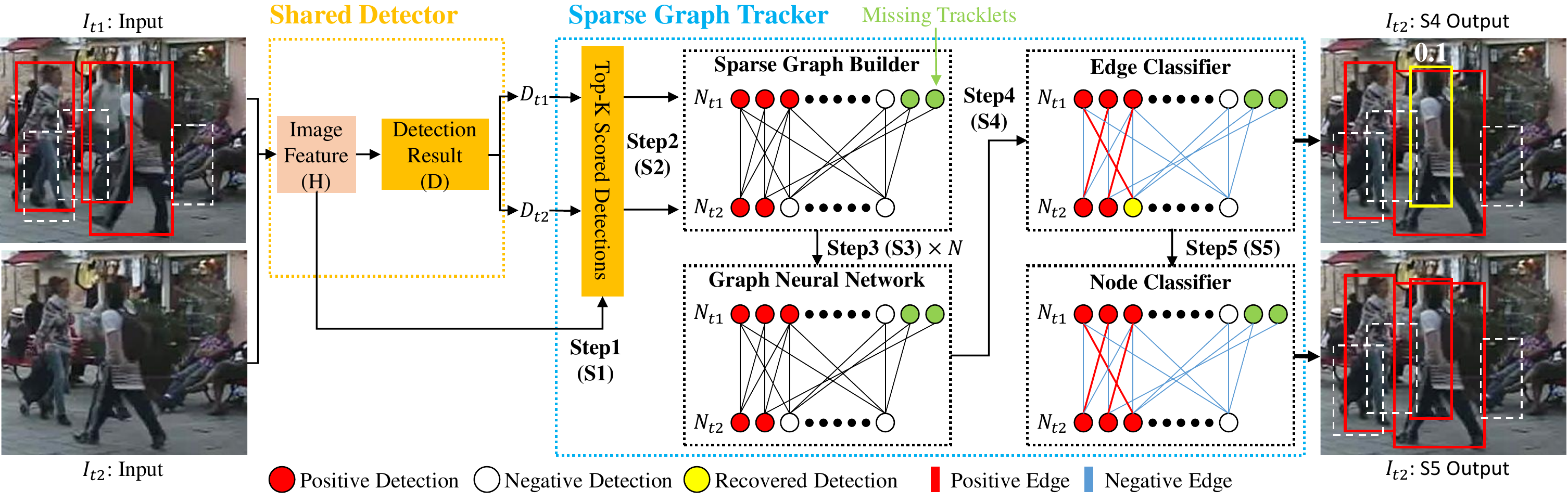}
    \caption{Overview of detection recovery in the inference pipeline of SGT.
    \emph{(S1)} Top-$K$ scored detections and their features are extracted from $I_{t1}$ and $I_{t2}$. The red boxes indicate positive detection that contains ID.
    \emph{(S2)} A sparse graph is built, where a node, $n^i_T \in N_T$ ($i \in [1, K]$), is a detection of frame $T=\{t1, t2\}$ and an edge ($e_{i,j}$) is a connection between $n^i_{t1}$ and $n^j_{t2}$. The green nodes are the tracklets that are missed until $t1$ and they are appended to $N_{t1}$. The red nodes indicate the positive detection that has its assigned ID.
    Two nodes at $N_{t2}$ are red since their detection scores are above $\tau_{init}$, so new ID can be assigned to them.
    \emph{(S3)} GNN updates the features of nodes and edges to become higher-order by aggregating neighboring features.
    \emph{(S4)} The edge score of the red line (a positive edge) is above the edge threshold ($\tau_E$) while the green line represents a negative edge. The yellow node ($n^3_{t2}$) is an example of detection recovery. It was previously negative detection due to its low score, but it becomes a positive detection with the help of a positive edge.
    \emph{(S5)} The recovered detection ($n^3_{t2}$) in S4 is verified by the node score. If the node score is below the node threshold ($\tau_N$), it is regarded as a false positive and is filtered out. Otherwise, the node is recovered and hence can be successfully detected which is shown by the yellow node becoming red.
    }
    \label{fig:overview}
\end{figure*}

\subsection{Online Detection Recovery}

In the TBD framework, detections to be tracked are decided based on certain threshold. 
Two detection threshold values are commonly used in online MOT methods~\cite{jde, fairmot, cstrack}: $\tau_{init}$ and $\tau_D$ for initializing unmatched detections as new tracklets and choosing tracking candidates, respectively. 
Due to a high value of $\tau_D$ (\eg,~0.4), some low-scored true detections are not included in the tracking candidates. 
In ByteTrack~\cite{bytetrack}, an extra association stage is deployed that the unmatched tracklets are matched with low-scored detections using IoU score.
However, a new detection threshold, $\tau_{D_{low}}$ (\eg,~0.2), is introduced for selecting the low-scored detections as the candidates, and this is a critical value deciding the trade-off between FP and FN. 
OMC~\cite{omcnet} introduces an extra stage before the association step to complement missed detections which may not be detected due to the low confidence score.

\section{Sparse Graph Tracker}
\label{sec:sgt}

\subsection{Overall Architecture}
\label{subsec:overview}

%% Use \mathit{conf}_{...} instead of conf_{}
%% P1. Overview of SGT 
%% Preliminary step & Step1
Figure~\ref{fig:overview} shows the architecture of SGT. While various image backbones and object detectors can be flexibly adopted to SGT, our main experiment is based on CenterNet~\cite{centernet} with a variant of DLA-34 backbone~\cite{dla} as same as our baseline, FairMOT~\cite{fairmot}. Following~\cite{fairmot}, we modify CenterNet that the box size predictor outputs left, right, top, and bottom sizes $(s_l, s_r, s_t, s_b)$ from a center point of an object instead of the width and height. CenterNet is a point-based detector that predicts object at every pixel of a feature map. The score head's output is denoted as $B_{score} \in \mathbb{R}^{H_h\times H_w\times 1}$, where $H_h$ and $H_w$ are height and width of the feature map. The output from the size head is denoted as $B_{size} \in \mathbb{R}^{H_h\times H_w\times 4}$. The offset head adjusts the center coordinates of objects using $B_{off} \in \mathbb{R}^{H_h\times H_w\times 2}$. At frame $T$, CenterNet outputs detections $D_T = {(S_T,\, B_T)}$, where $S_T$ is the detection score ($B_{score}$) and $B_T \in \mathbb{R}^{H_h\times H_w \times 4}$ is top-left and bottom-right coordinates.

%% Details of Step2 and stating how our contribution achieved thr a sparse graph
%% How to build a graph - node and edge features
\noindent \textbf{Sparse graph builder} takes top-$K$ scored detections from each frame ($\mathit{I}_{t1}$ and $\mathit{I}_{t2}$) and sets them as the nodes of a graph ($N_{t1}$ and $N_{t2}$). In the inference phase, the previous timestep's $N_{t2}$ will be the current timestep's $N_{t1}$. We sparsely connect $N_{t1}$ and $N_{t2}$ only if they are close in either Euclidean or feature space. Specifically, $n^i_{t1} \in N_{t1}$ is connected to $N_{t2}$ with three criteria: 1) small distance between their center coordinates; 2) high cosine similarity between their features; 3) high IoU score. For each criterion, the given number of $N_{t2}$ (\eg,~10) are selected to be connected to $n^i_{t1}$ without duplicates. The connection is bidirectional so that both $N_{t1}$ and $N_{t2}$ update their features. The visual features of the detections and relational features are used as the features of nodes ($V$) and edges ($E$), respectively.

%% Why do we use top-k detection, instead of using lower threshold?
To include low-scored detections for tracking, a low threshold value can be an alternative to top-$K$. Although it can also achieve good performance if it is small enough as shown in the supplementary material, such detection threshold value is sensitive to the detector's score distribution. As a result, careful calibration is required for different detectors and datasets. In contrast, top-$K$ method is robust to such issues as it is not affected by the score distribution. Since $K$ is the maximum number of objects that the model can track, we set $K$ to be sufficiently larger than the maximum number of people in the dataset (\eg,~100 in MOT16/17; 300 in MOT20). In Table~\ref{table:topk_robustness}, we experimentally show the robustness of $K$ values.

%% About long-term association handling - we can achieve without additional motion predictor (e.g. Kalman filter)
Some tracklets are failed to track for a while when they are invisible due to full occlusion. These missing tracklets are stored for a period of $\mathit{age}_{max}$ and appended to $N_{t1}$. Although existing MOT works~\cite{IOUTracker, jde, fairmot, corrtracker} apply a motion predictor (\eg,~Kalman filter~\cite{kalman_filter}) for predicting the possible location of missing tracklets, SGT can perform long-term association without a motion predictor. Here, we store the tracklets whose length is longer than $\mathit{age}_{min}$ to prevent false positive cases.

\begin{figure}[t!]
    \centering
    \includegraphics[width=\linewidth]{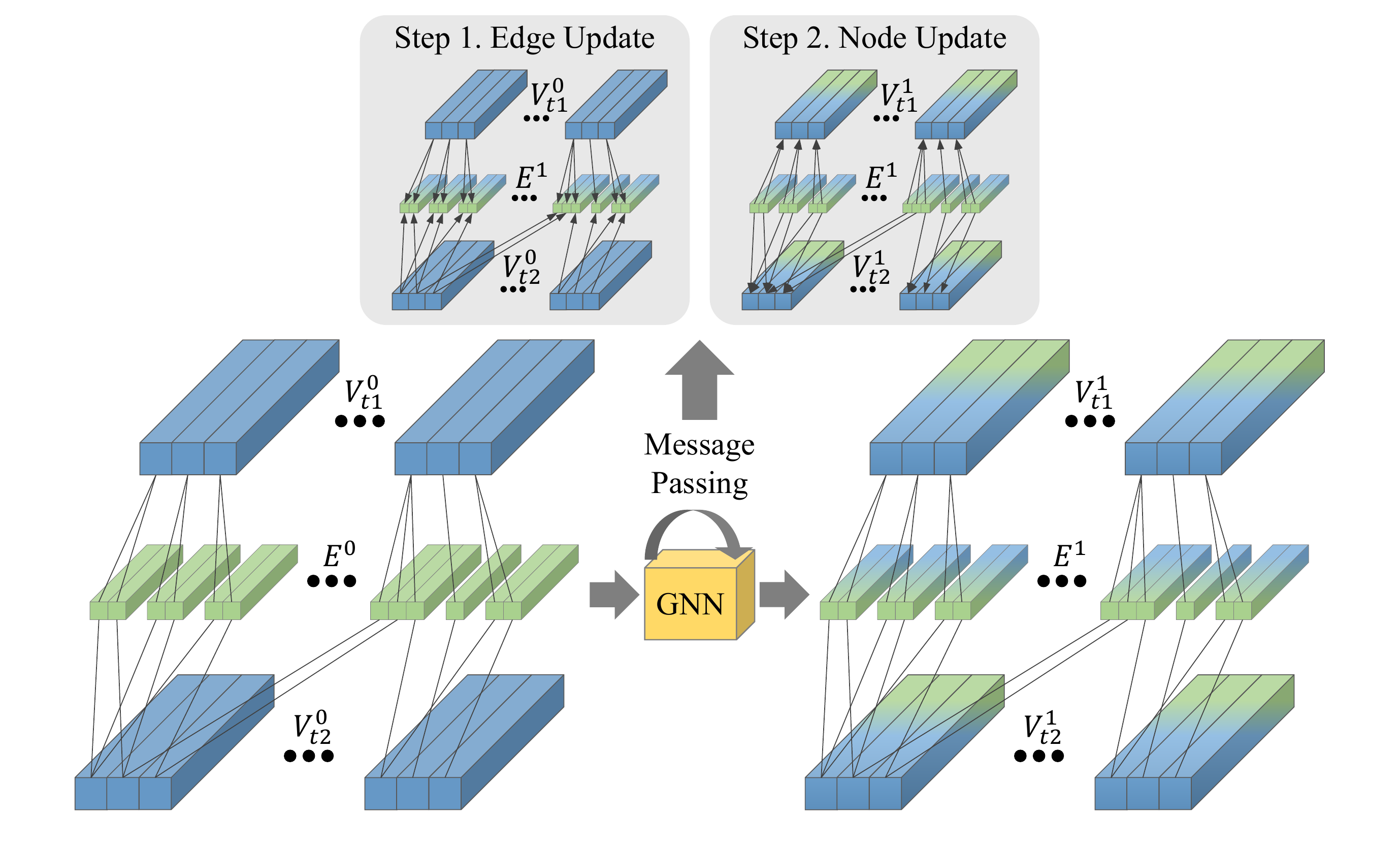}
    \caption{
    Illustration of message passing in a GNN. Initial edge features, $E^0$, are updated to $E^1$ containing the features of two connected nodes, $V^0_{t1}$ and $V^0_{t2}$. Initial node features, $V^0$, are then updated to $V^1$ containing the features of connected nodes, $V^0$, and the updated edges features, $E^1$. For simplicity, we omit bidirectional connections and show only few edges.
    % GNN's message passing process updates node and edge features in two steps. Initial edge features, $E^0$, are updated to $E^1$ containing the features of two connected nodes, $V^0_{t1}$ and $V^0_{t2}$. Initial node features, $V^0$, are then updated to $V^1$ containing the features of connected nodes, $V^0$ and updated edges features, $E^1$. We omit bidirectional connections and show only few edges for simplicity.
    }
    \label{fig:GNN}
\end{figure}

%% P3. GNN
\noindent \textbf{Graph neural network} updates features of nodes ($V$) and edges ($E$) in a graph through the message passing process, as described in Figure~\ref{fig:GNN}, that propagates the features to the neighboring nodes and edges and then aggregates them.
By iterating this process, $V$ now contains the features of both the neighboring nodes and edges, and $E$ indirectly aggregates the features of other edges which are connected to the same node.
While the initial edge features represent the pairwise relation of two detections, the iteration of the process allows the updated edge features to represent the higher-order (multi-hop) relation that also considers neighboring detections.
Section~\ref{subsec:sparse_graph_tracker} presents more details.

%% Step4 - edge classifier - learns object-level spatio-temporal consistency
\noindent \textbf{Edge classifier} is a FC layer that predicts the edge score ($ES$) from the updated edge features. 
The edge score is the probability that the connected detections at $t1$ and $t2$ refer to the same object.
Since $n^i_{t1}$ is connected to many nodes at $t2$, we use the Hungarian algorithm \cite{hungarian_matching} for optimal matching based on the edge score matrix. 
As a result, $n^i_{t1}$ has only one edge score which is optimally assigned. 
Then, the edge threshold ($\tau_E$) is used for deciding a positive or negative edge. 
The yellow box shown in Figure~\ref{fig:overview} is the recovered detection that $n^3_{t2}$ is negative due to its low detection score, but its connected node, $n^1_{t1}$, and edge ($e_{1,3}$) are positive.

%% Step5 - node classifier
\noindent \textbf{Node classifier} is a FC layer that prevents incorrect detection recovery by predicting the node score ($NS$) from the updated node features.
If the recovered detection's node score is below the node threshold ($\tau_N$), we decide not to recover it, thus the node stays negative.
Otherwise, we confirm recovery of the missed detection and the node becomes positive as shown by $n^3_{t2}$ in Figure~\ref{fig:overview}.

\subsection{Graph Construction and Update}
\label{subsec:sparse_graph_tracker}

This section explains design of the node and edge features in SGT. Note that a FC block refers to a stack of FC layer, layer normalization~\cite{layernorm}, and ReLU function 
% Designing the node and edge features is crucial for graphs.
% Here, we use a FC layer, layer normalization~\cite{layernorm} and ReLU activation function as a basic FC block.

\noindent \textbf{Initial node features.} 
Contrary to the graph-based MOT works using reID features of detected objects~\cite{tmpnn, strn, gnn3dmot}, SGT exploits the image backbone's visual features ($H$) which are shared for detection and jointly trained.

\noindent \textbf{Initial edge features.}
Edge feature are denoted as $e^l_{i, j}$, where $i$ and $j$ are the starting and ending node indices respectively, and $l$ indicates iteration.
Inspired by MPNTrack~\cite{tmpnn}, SGT initializes high-dimensional edge features as Eq.~\ref{eq:edge_attr}.
\begin{equation}
    \label{eq:edge_attr}
    \resizebox{0.42\textwidth}{!}{$ 
    e^0_{i,\, j} = f_{enc}\left( \left[ 
        x_i-x_j,\, y_i-y_j,\, \log({\frac{w_{i}}{w_{j}}}),\, \log({\frac{h_{i}}{h_{j}}}), \, {IoU}_{i,\, j},\, Sim_{i,\, j}
    \right] \right),
    $}
\end{equation}
where $[\cdot]$ is concatenation operator, $x$ and $y$ are the center coordinates, $h$ and $w$ are the height and width of a bounding box, $Sim$ is cosine similarity, and $f_{enc}$ refers to two FC blocks.
As the initialized edge features are direction-aware, two edges connecting the same nodes but reversely will have different features considering different relations (\eg,~$t1 \rightarrow t2$ and $t2 \rightarrow t1$).
$V_{t1}$ and $V_{t2}$ are updated on two different MLPs with these different edge features.
After updating in GNN, these bidirectional edge features are averaged to predict a single edge score.

\noindent \textbf{Initial graph}, shown in the left of Figure~\ref{fig:GNN}, is denoted by $G^0=\{ V^0,\, E^0 \}$, where $E^0 = \{e^0_{i, j}|\, 1 \le i,j \le 2K+|V_{miss}| \}$ is a set of initial edge features and $V^0=V^0_{t1} \bigcup V^0_{t2} \bigcup V^0_{miss}$ is a set of initial node features at $t1$, $t2$, and missing tracklets. 

\noindent \textbf{Update in node and edge features.} 
Figure~\ref{fig:GNN} describes two steps to update the features of nodes and edges during the message passing process in GNN.
The initial edge features $e^0_{i,\, j}$, shown in the left side of the graph, are pairwise relational features considering only the two connected nodes at $t1$ and $t2$ (direction $i \rightarrow j$).

In Step 1 of Figure~\ref{fig:GNN}, the edge features are updated as Eq.~\ref{eq:edge_update}.
\begin{equation}
    \label{eq:edge_update}
    % \resizebox{0.9\textwidth}{!}{$
    e^{l}_{i,\, j} = f_e\left( \left[ v^{l-1}_{i},\, v^{l-1}_{j},\, e^0_{i,\, j},\,  e^{l-1}_{i,\, j}  \right] \right),
    % $}
\end{equation}
where $f_e$ refers to two FC blocks, $l$ is the number of iterations ($l \in [1,N_{iter}]$), $v_i$ is the features of node $i$, and $v_i^{l-1}$ indicates the node features of the previous iteration.
Therefore, the current state of the two connected nodes, initial and current edge features are concatenated and passed to $f_e$ to update edge features as $e^{l}_{i,\, j}$.
Initial edge features ($e^0_{i,j}$) are concatenated every iteration to prevent the over-smoothing issue in GNN \cite{oversmoothing_issue}. 
Although we use the shared MLPs ($f_e$) for the edges of two different directions, the edge features of the opposite direction may not be the same since their edge features are encoded in a direction-aware manner.

In Step 2 of Figure~\ref{fig:GNN}, node $j$ aggregates the features of the connected nodes and edges as Eq.~\ref{eq:node_update}.
\begin{equation}
    \label{eq:node_update}
    v^{l}_{j} = f_{v_{out}}\left( \frac{1}{|E^l_{:, j}|} \sum_{i} f_{v_{enc}} \left( \left[ v^{l-1}_{i},\, e^{l}_{i,\,j}  \right] \right) \right),
\end{equation}
where $f_{v_{out}}$ is an FC block, $|E^l_{:,j}|$ is the number of edges connected to the node $j$, $f_{v_{enc}}$ refers to two FC blocks, $e^l_{i\, j}$ is the updated edge features in Step 1 (Eq.~\ref{eq:edge_update}) and $v^{l-1}_i$ is the features of starting node.
We suppose the index of $N_{t2}$ is from $1$ to $K$ and $N_{t1}$ is from $K+1$ to $2K+|V_{miss}|$.
When $i>j$, $e_{i\, j}$ is the edge features with direction of $t1 \rightarrow t2$. 
Thus, message passing is from $t1$ to $t2$ and $V_{t2}$ are updated.
As our edge features are direction-aware, we use different $f_{v_{enc}}$ for message passing $t1 \rightarrow t2$ and $t2 \rightarrow t1$.

\subsection{Training and Inference Techniques}
\label{subsec:train}

SGT is trained by the sum of the detection loss ($\mathcal{L}_{D}$) and the association loss ($\mathcal{L}_{A}$).

\noindent \textbf{Detection loss.} 
Since we adopt CenterNet \cite{centernet} as a detector, we follow \cite{centernet} to compute the detection loss which is the weighted sum of losses from three heads as Eq.~\ref{eq:det_loss}.
\begin{equation}
    \label{eq:det_loss}
    \mathcal{L}_{D} = \mathcal{L}_{score} + w_{size}  \mathcal{L}_{size} + w_{off}  \mathcal{L}_{off}
\end{equation}
%% Losses of size and offset 
The size head outputs $B_{size}$ composed of  $(s_l, s_r, s_t, s_b)$.
The offset head outputs $B_{off}$ which is the quantization error of the center coordinates caused by the stride of feature map (\eg,~4).
For each ground-truth (GT) object $\hat{b}^i=(\hat{x}^i_l, \hat{y}^i_t, \hat{x}^i_r, \hat{y}^i_b)$, GT size $ \hat{b}^i_{size}=(\hat{s}^i_l, \hat{s}^i_r, \hat{s}^i_t, \hat{s}^i_b)$ is computed by the difference between center coordinates $(\hat{c}^i_x, \hat{c}^i_y) = (\frac{\hat{x}^i_l + \hat{x}^i_r}{2}, \frac{\hat{y}^i_t + \hat{y}^i_b}{2})$ and $\hat{b}^i$.
Each GT size $\hat{b}^i_{size}$ is assigned to the prediction $b^{xy}_{size} \in B_{size}$, where $(x,y)=(\lfloor \frac{\hat{c}^i_x}{4} \rfloor, \lfloor \frac{\hat{c}^i_y}{4} \rfloor)$.
Each GT offset $(\hat{o}_x, \hat{o}_y) = (\frac{\hat{c}^i_x}{4} - \lfloor \frac{\hat{c}^i_x}{4} \rfloor, \frac{\hat{c}^i_y}{4} - \lfloor \frac{\hat{c}^i_y}{4} \rfloor)$ is assigned to the prediction $b^{xy}_{off}$.
Then, $l_1$ loss is used to compute $\mathcal{L}_{size}$ and $\mathcal{L}_{off}$.
%% Loss of heatmap
For training the score head, GT heatmap $M^{xy} \in \mathbb{R}^{H_h\times H_w\times 1}$ is generated by the Gaussian kernel as Eq.~\ref{eq:heatmap_gen}.
\begin{equation}
    \label{eq:heatmap_gen}
    \resizebox{0.34\textwidth}{!}{$
    M^{xy} =
    \sum\limits_{i=1}^{N_{D}}
    \exp(-\frac
    % {(x - c^i_x)^2) + (y - c^i_y)^2)}
    {(x - \lfloor \frac{\hat{c}^i_x}{4} \rfloor)^2) + (y - \lfloor \frac{\hat{c}^i_y}{4} \rfloor)^2)}
    {2\sigma^2_d}
    ),
    $}
\end{equation}
where $N_{D}$ is the number of GT object and $\sigma_d$ is computed by width and height of each object \cite{cornernet}.
$\mathcal{L}_{score}$ is computed as the pixel-wise logistic regression with the penalty-reduced focal loss \cite{focal_loss}.

\noindent \textbf{Association loss.}
Our association loss is the weighted sum of the edge and node classification losses as Eq.~\ref{eq:asso_loss}.
\begin{equation}
    \label{eq:asso_loss}
    \mathcal{L}_{A} = w_{edge} \mathcal{L}_{edge} + w_{node} \mathcal{L}_{node}
\end{equation}
In SGT, the edge and node classifiers output the edge and node scores ($ES$ and $NS$), respectively. 
$\mathcal{L}_{edge}$ and $\mathcal{L}_{node}$ are computed on these scores with the focal loss \cite{focal_loss}.
Since it is difficult to assign GT labels to the edges connecting the background patches, we exclude them in $\mathcal{L}_{edge}$ as Eq.~\ref{eq:edge_loss}. 
\begin{equation}
    \label{eq:edge_loss}
    \resizebox{0.46\textwidth}{!}{$
    \begin{multlined}
    \mathcal{L}_{edge}=
    \frac{1}{N_{E^{+}}}
    \sum\limits_{e_{i,j} \in E}
    \begin{cases}
        \text{FL}(\mathit{ES}_{i,j}, {ey}_{i,j}), & \text{if } \mathit{ny}_i=1 \text{ or } \mathit{ny}_j=1; \\
        % \text{FL}(\mathit{ES}_{i,j}, {ey}_{i,j}), & \text{if } \mathit{ny}=1; \\
        0 & \text{otherwise}, 
    \end{cases} 
    \end{multlined}
    $}
\end{equation}
where $N_{E^{+}}$ is the number of GT edges which at least one of the endpoints is positive, $E$ is a set of edges in $G$, FL is the focal loss, edge in direction of $t1 \rightarrow t2$, $\mathit{ey}_{i,j}$ is the GT label of edge connecting the nodes $n_i$ and $n_j$, and $\mathit{ny}_{i}$ is the GT label of $n_{i}$.
We compute $\mathcal{L}_{node}$ only on the node scores at $t2$ as Eq.~\ref{eq:node_loss}. 
\begin{equation}
    \label{eq:node_loss}
    \resizebox{0.26\textwidth}{!}{$
    \mathcal{L}_{node} = \frac{1}{N_{N^{+}_{t2}}}
                % \sum\limits_{j=1}^{2K+|V_{miss}|}
                \sum\limits_{n_j \in N_{t2}}
                \text{FL}(\mathit{NS}_{j}, \mathit{ny}_j)
                ,
    $}
\end{equation}
where $N_{N^{+}_{t2}}$ is the number of GT positive nodes at $t2$.
We output zero when $N_{E^{+}}=0$ or $N_{N^{+}_{t2}}=0$.

\noindent \textbf{Node and edge label assignment} is an essential step for computing the association loss.
While existing GNN-based tracker~\cite{tmpnn} trains its matching network using GT objects, we introduce a novel training technique using pseudo labels to effectively train the edge and node classifiers in a single step with a detector and a shared backbone network.
The top-$K$ detections are optimally matched with the GT objects based on their IoU score matrix and Hungarian algorithm~\cite{hungarian_matching}, and IDs of the GTs are assigned to the matched detections. To prevent the misallocation of GT ID, the assigned IDs are filtered out if IoU of their matching is lower than the threshold (\eg,~0.5). This step is repeated for $N_{t1}$ and $N_{t2}$ to assign ($\mathit{ny}_i$ and $\mathit{ny}_j$).
Finally, the GT edge label ($\mathit{ey}_{i,j}$) is assigned to the edges by matching the IDs of nodes. 
An edge is labeled as 1 if the two connected nodes have the same GT ID, and 0 otherwise.

\noindent \textbf{Adaptive feature smoothing} (AdapFS) is a novel inference technique for the proposed detection recovery framework. Following JDE~\cite{jde}, recent online TBD models update appearance features of tracklets in an exponential moving average manner as $\mathit{emb}_{t2}^{trk} = \alpha \times \mathit{emb}_{t1}^{trk} + (1-\alpha) \times \mathit{emb}_{t2}^{det}$. The features of tracklets are updated by adding the features of new detections with the fixed weight, $\alpha$. However, low-scored recovered objects have unreliable appearance features since they may suffer from occlusion or blur. Thus, we incorporate adaptive weight computed by the object scores ($S_T$) of matched tracklets and detections as Eq.~\ref{eq:adap_feat_smooth}.
\begin{equation}
    \label{eq:adap_feat_smooth}
    \resizebox{0.4\textwidth}{!}{$
    \mathit{emb}_{t2}^{trk} = \mathit{emb}_{t1}^{trk} \times \frac{S_{t1}}{S_{t1} + S_{t2}} + \mathit{emb}_{t2}^{det} \times \frac{S_{t2}}{S_{t1} + S_{t2}}
    $}
\end{equation}

\section{Experiments}
\label{sec:experiments}

\subsection{Datasets and Implementation Details}
\label{sec:dataset_impl_details}

\noindent \textbf{Datasets.} 
We train and evaluate the proposed method using MOT16/17/20 and HiEve Challenge datasets~\cite{mot16, mot20, hieve} which target pedestrian tracking. MOT20 and HiEve are complex datasets composed of crowded scenes. On each frame, MOT20 has 170 people on average compared to MOT17 containing 30 people. Due to the small size of the MOT datasets, JDE~\cite{jde} introduces the pedestrian detection and reID datasets~\cite{eth, cityperson, caltech, cuhk-sysu, prw} for training. FairMOT~\cite{fairmot} further exploits extra pedestrian detection dataset, CrowdHuman~\cite{crowdhuman}. We only use CrowdHuman as an extra training dataset to achieve competitive performance. Since CrowdHuman does not have ID labels and is not a video dataset, we assign a unique ID to every object and randomly warp an image to generate a pair of consecutive frames ($I_{t1}$--$I_{t2}$).

\noindent \textbf{Implementation details.} 
We use CenterNet~\cite{centernet} pretrained on COCO object detection dataset~\cite{mscoco_detection_dataset} to initialize SGT's detector. For fair comparison with~\cite{fairmot, corrtracker, relationtrack, gsdt}, we use the image size of $1088 \times 608$ and the feature map size ($H_w \times H_h$) of $272 \times 152$. Two consecutive frames are randomly sampled in the interval of $[1, 30]$. Following \cite{fairmot}, random flip, warping and color jittering are selected as data augmentation. The same augmentation is applied to a pair of images. We use Adam optimizer~\cite{adam_opt} with a batch size of 12 and initial learning rate (lr) of $2e^{-4}$ which drops to $2e^{-5}$. There are 60 training epochs and lr is dropped at 50. For training, we use 1 for $w_{off}$, 0.1 for $w_{size}, w_{edge}$, and 10 for $w_{node}$. For inference, we use 0.5, 0.4 and 0.4 as $\tau_{init}$, $\tau_E$ and $\tau_N$, respectively. These values are chosen empirically.

\subsection{MOT Challenge Evaluation Results}
We submit our result to the MOT16/17/20 Challenge test server and compare it with the recent online MOT models as shown in Table~\ref{table:sgt_mot_result}. Note that the methods using tracklet interpolation as post-processing (\eg,~ByteTrack~\cite{bytetrack}) are excluded to satisfy \emph{online setting}. Visualization results are provided in the supplementary material.

\noindent \textbf{Evaluation metrics.} 
We use the standard evaluation metrics for 2D MOT~\cite{mot_metrics}: Multi-object Tracking Accuracy (MOTA), ID F1 Score (IDF1), False Negative (FN), False Positive (FP), and Identity Switch (IDS)~\cite{idswitch}. While MOTA is computed by FP, FN, and IDS, and thus focuses on the detection performance, IDF1~\cite{idf1} is a metric focused on tracking performance. Also, mostly tracked targets (MT) and mostly lost targets (ML) represent the ratio of GT trajectories covered by a track hypothesis for at least 80\% and at most 20\% of their respective life span, respectively.

\begin{table}[t!]
\caption{
Evaluation results of ours and recent online JDT models on the MOT16/17/20 benchmarks (private detection). OMC--F~\cite{omcnet} applies its method on FairMOT~\cite{fairmot}. For each metric, the best is bolded and the second best is underlined. The values not provided are filled by ``-''. \textbf{\dag}~indicates no extra training dataset.
}
\label{table:sgt_mot_result}
\centering
\resizebox{\linewidth}{!}{
\begin{tabular}{@{}l|ccccccc@{}}
    \toprule
    Method & MOTA$\uparrow$ & IDF1$\uparrow$ & MT$\uparrow$ & ML$\downarrow$ & FP$\downarrow$ & FN$\downarrow$ & IDS$\downarrow$ \\
    \midrule
    \midrule
    \multicolumn{8}{c}{MOT16 \cite{mot16}} \\
    \midrule
        QDTrack \cite{quasi-dense} \textbf{\dag}  & 69.8 & 67.1 & 41.6 & 19.8 & 9861 & 44050 & 1097 \\
        TraDes \cite{trades} & 70.1 & 64.7 & 37.3 & 20.0 & \textbf{8091} & 45210 & 1144 \\
        CSTrack \cite{cstrack} \textbf{\dag} & 71.3 & 68.6 & - & - & - & - & 1356 \\
        \textbf{SGT (Ours)} \textbf{\dag} & 74.1 & 71.0 & 43.6 & 15.8 & 9784 & 35946 & 1528 \\
        GSDT \cite{gsdt} & 74.5 & 68.1 & 41.2 & 17.3 & \underline{8913} & 36428 & 1229 \\
        FairMOT \cite{fairmot} & 74.9 & 72.8 & 44.7 & 15.9 & - & - & 1074 \\
        CSTrack \cite{cstrack} & 75.6 & 73.3 & 42.8 & 16.5 & 9646 & 33777 & 1121 \\
        RelationTrack \cite{relationtrack} & 75.6 & \textbf{75.8} & 43.1 & 21.5 & 9786 & 34214 & \textbf{448} \\
        OMC \cite{omcnet} & 76.4 & 74.1 & 46.1 & \underline{13.3} & 10821 & 31044 & - \\
        CorrTracker \cite{corrtracker} & \underline{76.6} & \underline{74.3} & \underline{47.8} & \underline{13.3} & 10860 & \underline{30756} & \underline{979} \\
        \textbf{SGT (Ours)} & \textbf{76.8} & 73.5 & \textbf{49.3} & \textbf{10.5} & 10695 & \textbf{30394} & 1276 \\
    \midrule
    \multicolumn{8}{c}{MOT17 \cite{mot16}} \\
    \midrule
        CTracker \cite{chainedtracker} \textbf{\dag} & 66.6 & 57.4 & 32.2 & 24.2 & 22284 & 160491 & 5529 \\
        CenterTrack \cite{centertrack} & 67.8 & 64.7 & 34.6 & 24.6 & \textbf{18489} & 160332 & \underline{3039} \\
        QDTrack \cite{quasi-dense} \dag & 68.7 & 66.3 & 40.6 & 21.9 & 26598 & 146643 & 3378 \\
        TraDes \cite{trades} & 69.1 & 63.9 & 36.4 & 21.5 & \underline{20892} & 150060 & 3555 \\
        FairMOT \cite{fairmot} \dag & 69.8 & 69.9 & - & - & - & - & 3996 \\
        SOTMOT \cite{sotmot} & 71.0 & 71.9 & 42.7 & 15.3 & 39537 & 118983 & 5184 \\
        GSDT \cite{gsdt} & 73.2 & 66.5 & 41.7 & 17.5 & 26397 & 120666 & 3891 \\
        \textbf{SGT (Ours)} \dag & 73.2 & 70.2 & 42.0 & 17.7 & 25332 & 121155 & 4809 \\
        FairMOT \cite{fairmot} & 73.7 & 72.3 & 43.2 & 17.3 & 27507 & 117477 & 3303 \\
        RelationTrack \cite{relationtrack} & 73.8 & \textbf{74.7} & 41.7 & 23.2 & 27999 & 118623 & \textbf{1374} \\
        TransTrack \cite{transtrack} & 74.5 & 63.9 & 46.8 & \textbf{11.3} & 28323 & 112137 & 3663 \\
        OMC--F \cite{omcnet} & 74.7 & \underline{73.8} & 44.3 & 15.4 & 30162 & 108556 & - \\
        CSTrack \cite{cstrack} & 74.9 & 72.3 & 41.5 & 17.5 & 23847 & 114303 & 3567 \\
        OMC \cite{omcnet} & 76.3 & \underline{73.8} & 44.7 & 13.6 & 28894 & \underline{101022} & - \\
        \textbf{SGT (Ours)} & \underline{76.4} & 72.8 & \textbf{48.0} & \underline{11.7} & 25974 & 102885 & 4101 \\
        CorrTracker \cite{corrtracker} & \textbf{76.5} & 73.6 & \underline{47.6} & 12.7 & 29808 & \textbf{99510} & 3369 \\
    \midrule
    \multicolumn{8}{c}{MOT20 \cite{mot20}} \\
    \midrule
        FairMOT \cite{fairmot} & 61.8 & 67.3 & \textbf{68.8} & \textbf{7.6} & 103440 & \textbf{88901} & 5243\\
        TransTrack \cite{transtrack} & 64.5 & 59.2 & 49.1 & 13.6 & 28566 & 151377 & 3565 \\
        \textbf{SGT (Ours)} \dag & 64.5 & 62.7 & 62.7 & 10.2 & 67352 & 111201 & 4909 \\
        CorrTracker \cite{corrtracker} & 65.2 & 69.1 & \underline{66.4} & \underline{8.9} & 79429 & \underline{95855} & 5183 \\
        CSTrack \cite{cstrack} & 66.6 & 68.6 & 50.4 & 15.5 & 25404 & 144358 & 3196 \\
        GSDT \cite{gsdt} & 67.1 & 67.5 & 53.1 & 13.2 & 31913 & 135409 & \underline{3131} \\
        RelationTrack \cite{relationtrack} & 67.2 & 70.5 & 62.2 & \underline{8.9} & 61134 & 104597 & 4243 \\
        SOTMOT \cite{sotmot} & 68.6 & \textbf{71.4} & 64.9 & 9,7 & 57064 & 101154 & 4209 \\
        OMC \cite{omcnet} & \underline{70.7} & 67.8 & 56.6 & 13.3 & \textbf{22689} & 125039 & - \\
        \textbf{SGT (Ours)} & \textbf{72.8} & \underline{70.6} & 64.3 & 12.7 & \underline{25161} & 112963 & \textbf{2474} \\
    \bottomrule
\end{tabular}
}
\end{table}

%% mot w/o extra training dataset

% \begin{table}[t!]
% % \setlength{\tabcolsep}{3pt} % Default value: 6pt
% \caption{
% Evaluation results of ours and recent online JDT models on the MOT/17/20 benchmarks (private detection) without extra training datasets. 
% }
% \label{table:sgt_mot_result}
% \centering
% \resizebox{\linewidth}{!}{
% \begin{tabular}{@{}l|ccccccc@{}}
%     \toprule
%     Method & MOTA$\uparrow$ & IDF1$\uparrow$ & MT$\uparrow$ & ML$\downarrow$ & FP$\downarrow$ & FN$\downarrow$ & IDS$\downarrow$ \\
%     \midrule
%     \midrule
%     \multicolumn{8}{c}{MOT17 \cite{mot16}} \\
%     \midrule
%         CTracker \cite{chainedtracker} & 66.6 & 57.4 & 32.2 & 24.2 & 22284 & 160491 & 5529 \\
%         QDTrack \cite{quasi-dense} & 68.7 & 66.3 & 40.6 & 21.9 & 26598 & 146643 & 3378 \\
%         FairMOT \cite{fairmot} & 71.9 & 72.3 & - & - & 18801 & 136821 & 2655 \\
%         TransTrack \cite{transtrack} &  &  &  &  &  &  &  \\
%         \textbf{SGT (Ours)} & 73.2 & 70.2 & 42.0 & 17.7 & 25332 & 121155 & 4809 \\
%     \midrule
%     \multicolumn{8}{c}{MOT20 \cite{mot20}} \\
%     \midrule
%         FairMOT \cite{fairmot} & 62.3 & 66.7 & 60.7 & 9.8 & 78431 & 111488 & 5073 \\
%         TransTrack \cite{transtrack} & \\
%         \textbf{SGT (Ours)} & 64.5 & 62.7 & 62.7 & 10.2 & 67352 & 111201 & 4909 \\
%     \bottomrule
% \end{tabular}
% }
% \end{table}

\noindent \textbf{Evaluation results of MOT16/17.}
Without extra training datasets, MOTA of SGT is higher than CSTrack~\cite{cstrack} and FairMOT~\cite{fairmot} by about 3\%, and it is comparable with FairMOT trained with extra training datasets.
With CrowdHuman as an extra training dataset, SGT achieves the highest MOTA on MOT16/17 based on the best trade-off between FP and FN. 
The highest MT indicates that SGT generates stable and long-lasting tracklets which is attributed to the proposed detection recovery mechanism. 
Compared with the models based on the same detector \cite{centertrack, quasi-dense, trades, gsdt, fairmot, relationtrack, corrtracker, sotmot, corrtracker}, SGT outperforms all of them, except CorrTracker~\cite{corrtracker} which shows marginally higher MOTA by 0.1\% on MOT17.
When we compare SGT with OMC--F~\cite{omcnet} which is also specially designed to track low-scored detections on top of FairMOT~\cite{fairmot}, SGT shows lower FN with lower FP, demonstrating the superiority of our detection recovery mechanism over OMC~\cite{omcnet}. 
SGT achieves excellent tracking performance on MOT20, while it shows high IDS on MOT17. 
This is caused by non-human occluders (\eg,~vehicles) frequently appearing in MOT17. 
Since pedestrian is the only target class for training the detector, other non-human objects are not included in top-$K$ detections and for relational modeling as well. 
This leads to inferior tracking performance in MOT17.

\noindent \textbf{Evaluation results of MOT20.}
In MOT20, scenes are severely crowded and partial occlusion is dominant compared with MOT16/17. 
When the same detection threshold ($\tau_D$) is employed, existing methods suffer from missed detections caused by less confident detection output.
\cite{fairmot, corrtracker, relationtrack}~use lower detection threshold for MOT20, yet this results in high FP, IDS and low IDF1 since their pairwise relational features are not strong enough for correctly matching with a large number of tracking candidates. 
On the other hand, our higher-order relational features allows SGT to effectively address this problem and achieve state-of-the-art in MOTA as shown in Table~\ref{table:sgt_mot_result}. 
SGT surpasses CorrTracker~\cite{corrtracker} in MOTA by 7.6\% while it shows higher MOTA than SGT in MOT17.
SGT achieves better trade-off between FN and FP, and higher MOTA and IDF1 than OMC~\cite{omcnet} whose detection recovery method is applied to CSTrack~\cite{cstrack}. 
Although OMC exploits past frame as temporal cues to carefully select the low-scored detections, its matching is still limited by the pairwise relational features. 
In contrast, SGT performs matching using the higher-order relational features updated by GNNs, and thus, SGT outperforms OMC in spite of the low-scored detections simply selected by top-$K$ sampling. Table~\ref{table:gnn_iter} further supports the importance of higher-order relational features.

% \begin{figure}[t!] 
%     \centering
%     \includegraphics[width=0.5\textwidth]{pics/sgt/sgt_mot20-score-vis.png}
%     \caption{Detection recovery cases in the MOT20 dataset \cite{mot20}. We show the annotation of each prediction box in the format of ``\{id\}-\{detection score\}''. 
%     % The detection threshold value is 0.5 so the objects which score below 0.5 are recovered detections.
%     }
%     \label{fig:sgt_mot20-recovery}
% \end{figure}

% \noindent \textbf{Visualization of recovered detections.}
% Figure~\ref{fig:sgt_mot20-recovery} shows the examples of detection recovery in MOT20 test dataset. 
% In the first row, people labeled by the blue and brown bounding boxes are occluded each other.
% From frame \#34 to \#37, their detection scores are below the detection threshold and they were missed detections originally; however, SGT successfully recovers them by including low-scored detections in the association step,
% Figure~\ref{fig:sgt_mot20-recovery} also shows the detection recovery case of people labeled by the green and orange bounding boxes in the second row.

\noindent \textbf{Inference speed.}
We measure the inference speed in terms of frames-per-second (FPS) using a single V100 GPU. SGT runs at 23.0/23.0/19.9 FPS on MOT16/17/20, respectively. For fair comparison, we select the methods reporting FPS measured on the same GPU. CorrTracker~\cite{corrtracker} and TransTrack~\cite{transtrack} run at 14.8 and 10.0 FPS, respectively, on MOT17, while CorrTracker runs at 8.5 FPS on MOT20. SGT runs much faster than them on both MOT17/20 since SGT performs relational modeling sparsely at object-level by top-$K$ sampling of detections, compared with them densely modeling the relationship of features at pixel-level.

\subsection{HiEve Challenge Evaluation}
\label{subsec:hieve_challenge}

Table~\ref{table:hieve_result} compares ours and online JDT models on the HiEve Challenge. Without extra training datasets, SGT achieves 47.2 MOTA and 53.7 IDF1. Under the condition without extra datasets, SGT shows a large improvement in both MOTA and IDF1 compared to FairMOT~\cite{fairmot} and CenterTrack~\cite{centertrack}. SGT can even achieve comparable MOTA and higher IDF1 than CSTrack~\cite{cstrack} which uses extra training datasets following the MOT benchmarks. 

\begin{table}[t!]
\centering
\caption{
Evaluation results of ours and recent online JDT models on the HiEve benchmark (private detection). For each metric, the best is bolded and the second best is underlined. \textbf{\dag}~indicates no extra training dataset.
}
\label{table:hieve_result}
\resizebox{\linewidth}{!}{
\begin{tabular}{@{}l|ccccccc@{}}
    \toprule
    Method & MOTA$\uparrow$ & IDF1$\uparrow$ & MT$\uparrow$ & ML$\downarrow$ & FP$\downarrow$ & FN$\downarrow$ & IDS$\downarrow$ \\
    \midrule
        DeepSORT \cite{deepsort} \textbf{\dag} & 27.1 & 28.5 & 8.4 & 41.4 & 5894 & 42668 & 3122 \\
        JDE \cite{jde} \textbf{\dag} & 33.1 & 36.0 & 15.1 & \textbf{24.1} & 6318 & 43577 & 3747 \\
        FairMOT \cite{fairmot} \textbf{\dag} & 35.0 & 46.7 & 16.3 & 44.2 & 6523 & 37750 & \textbf{995} \\
        CenterTrack \cite{centertrack} \textbf{\dag} & 40.9 & 45.1 & 10.8 & 32.2 & \underline{3208} & 36414 & 1568 \\
        NewTracker \cite{trackrcnn} & 46.4 & 43.2 & \textbf{26.3} & 30.8 & 4667 & \textbf{30489} & 2133 \\
        \textbf{SGT (Ours)} \textbf{\dag} & \underline{47.2} & \textbf{53.7} & \underline{24.0} & \underline{28.8} & 4699 & \underline{30727} & \underline{1361} \\
        CSTrack \cite{cstrack} & \textbf{48.6} & \underline{51.4} & 20.4 & 33.5 & \textbf{2366} & 31933 & 1475 \\
    \bottomrule
\end{tabular}
}
\end{table}

\begin{figure}[t!]
    \centering
    \includegraphics[width=0.47\textwidth]{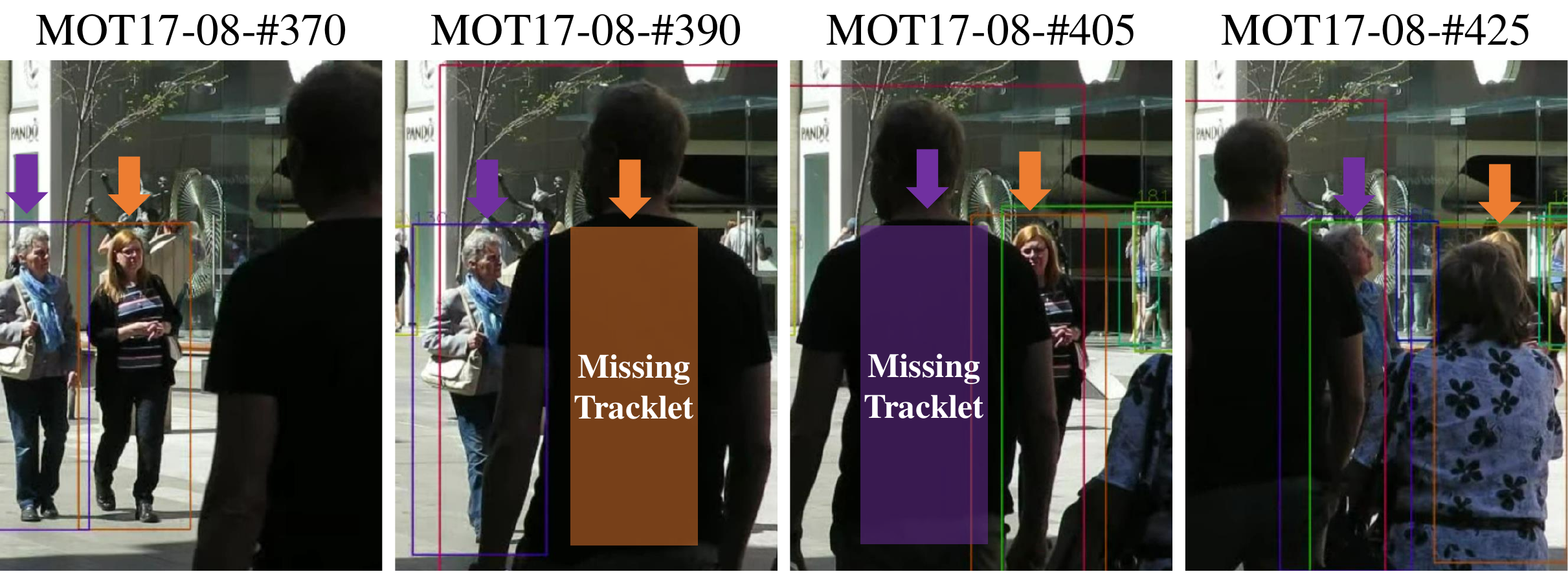}
    \caption{
    Illustration of long-term association in SGT without Kalman filter~\cite{kalman_filter}. Each color indicates the unique ID.
    }
    \label{fig:sgt_occlusion_longterm}
\end{figure}

\subsection{Ablation Experiments}
\label{subsec:ablation_study}
Ablation experiments are conducted by training model on the first half of the MOT17 train dataset and evaluating it on the rest. More analysis on detection recovery and ablation experiments can be found in the supplementary material.

\begin{table}[t!]
    \centering
    \caption{
    Ablation study of the detection recovery (DR) method. DR--GNN denotes DR by tracking using the edge features updated by GNNs. NC denotes the node classifier that  filtering out recovered detections using the node classifier. BG denotes that top-$K$ scored detections of $I_{t1}$ are used as $N_{t1}$.
    }
    \label{table:DR_ablation}
    \resizebox{\linewidth}{!}{
    \begin{tabular}{ @{} l | c c c | c c c c c c @{}}
        \toprule
        Model & DR--GNN & NC & BG & MOTA$\uparrow$ & IDF1$\uparrow$ & MT$\uparrow$ & FP$\downarrow$ & FN$\downarrow$ & IDS$\downarrow$\\
        \midrule
        \multirow{3}{*}{\begin{tabular}{@{} l} \textbf{SGT (Ours)} \end{tabular}}
         & \cmark & & & 70.7 & 73.3 & \textbf{49.9} & 3400 & \textbf{11794} & 619 \\ %ID-361
         & \cmark & \cmark & & 70.8 & 73.3 & 45.4 & \textbf{1952} & 13074 & 750 \\ % ID-360
         & \cmark & \cmark & \cmark & \textbf{71.3} & \textbf{73.8} & 46.6 & 2190 & 12742 & 588 \\ % ID-351
        \midrule
        FairMOT~\cite{fairmot} & & & & 69.6 & 72.5 & 44.0 & 2681 & 13341 & 414 \\ % ID-244
        BYTE~\cite{bytetrack} & & & & 69.7 & 73.3 & 47.5 & 3638 & 12347 & \textbf{400} \\ % ID-247
        \bottomrule
    \end{tabular}
    }
\end{table}

\noindent \textbf{Detection recovery.}
We conduct analysis of our architecture and comparison with another detection recovery method, BYTE~\cite{bytetrack}, and the results are shown in Table~\ref{table:DR_ablation}. Spatio-temporal relational modeling with top-$K$ detections of the previous frame benefits SGT to extract discriminative edge features. Otherwise, tracking performance is degraded with high IDS. In SGT, low-scored detections are also used for tracking and can be recovered, but FP recovery could be occurred. The node classifier verifies recovery cases with the updated node features and reduces FP. When we compare DR-GNN with BYTE~\cite{bytetrack} applied on FairMOT~\cite{fairmot}, DR-GNN achieves better trade-off between FP and FN, and consequently, higher MOTA. This demonstrates the effectiveness of edge features updated in the GNN.

\begin{table}[t]
    \centering
    \caption{Ablation study of training techniques. J and P denote joint training and pseudo labeling, respectively.}
    \label{table:train_tech_ablation}
    \resizebox{0.7\linewidth}{!}{
    \begin{tabular}{ @{} c c | c c c c c c @{} }
    \toprule
    P & J & MOTA$\uparrow$ & IDF1$\uparrow$ & MT$\uparrow$ & FP$\downarrow$ & FN$\downarrow$ & IDS$\downarrow$\\
    \midrule
     & & 38.5 & 55.8 & \textbf{54.3} & 21678 & \textbf{10136} & 1418  \\ % ID-353
    \cmark & & 69.1 & 69.7 & 47.2 & 3478 & 12360 & 847  \\ % ID-352
    \cmark & \cmark & \textbf{71.3} & \textbf{73.8} & 46.6 & \textbf{2190} & 12742 & \textbf{588} \\ % ID-351
    \bottomrule
    \end{tabular}}
\end{table}
\noindent \textbf{Training strategy.} 
According to Table~\ref{table:train_tech_ablation}, pseudo labeling based on top-$K$ detections is an important training technique for SGT. Both object--object and object--background pairs are included in edge labels by employing top-$K$ detections as pseudo labels. Training with object--background pairs as extra negative examples effectively supervises SGT not to false positively match. Jointly training detector and tracker leads to better performance, instead of using frozen backbone which is pretrained with detector.

% For validating effectiveness of joint training SGT and a detector, we 
% When SGT is jointly trained with a detector, it shows higher MOTA and IDF1 compared with SGT tra
% Instead of training SGT with the frozen detector which is trained , jointly trained SGT shows higher MOTA and IDF1.

% only mot17
\begin{table}[t!]
    \centering
    \caption{
    Effectiveness of adaptive feature smoothing (AdapFS).
    }
    \label{table:adapsf_ablation}
    \resizebox{\linewidth}{!}{
    \begin{tabular}{ @{} c c | c c c c c c @{}}
    \toprule
    Feature Soothing & Adaptive Weight & MOTA$\uparrow$ & IDF1$\uparrow$ & MT$\uparrow$ & FP$\downarrow$ & FN$\downarrow$ & IDS$\downarrow$ \\
    \midrule
     & & 71.3 & 73.1 & 46.9 & 2178 & 12724 & 590 \\ % ID-457
    \cmark & & 71.4 & 72.9 & 46.9 & 2170 & 12713 & 589 \\ % ID-456
    \cmark & \cmark & 71.3 & 73.8 & 46.6 & 2190 & 12742 & 588 \\ % ID-455
    \bottomrule
    \end{tabular}}
\end{table}

% % CH + mot17 version
% \begin{table}[t!]
%     \centering
%     \caption{
%     Effectiveness of adaptive feature smoothing (AdapFS). CrowdHuman~\cite{crowdhuman} is used for training SGT.
%     }
%     \label{table:adapsf_ablation}
%     \resizebox{\linewidth}{!}{
%     \begin{tabular}{ @{} c c | c c c c c c @{}}
%     \toprule
%     Feature Soothing & Adaptive Weight & MOTA$\uparrow$ & IDF1$\uparrow$ & MT$\uparrow$ & FP$\downarrow$ & FN$\downarrow$ & IDS$\downarrow$ \\
%     \midrule
%      & & 73.9 & 75.3 & 53.4 & 2533 & 11076 & 478 \\ % ID-330
%     \cmark & & 74.0 & 75.6 & 53.1 & 2556 & 11034 & 468 \\ % ID-329
%     \cmark & \cmark & 74.2 & 76.3 & 53.1 & 2514 & 10978 & 451 \\ % ID-328
%     \bottomrule
%     \end{tabular}}
% \end{table}

\noindent \textbf{Effectiveness of AdapFS.}
With the fixed weight, IDF1 is slightly decreased as shown in Table~\ref{table:adapsf_ablation}. On the other hand, IDF1 increases and tracking performance is improved based on our proposed adaptive feature smoothing in SGT. 

\begin{table}[t!]
    \centering
    \caption{
    Ablation study of the long-term association in SGT. The unit of $\mathit{age}_{max}$ is second while that of $\mathit{age}_{min}$ is frame.
    }
    \label{table:long_term}
    \resizebox{0.9\linewidth}{!}{
    \begin{tabular}{ @{} c c|c c c c c c @{}}
    \toprule
    $\mathit{age}_{max}$ & $\mathit{age}_{min}$ & MOTA$\uparrow$ & IDF1$\uparrow$ & MT$\uparrow$ & FP$\downarrow$ & FN$\downarrow$ & IDS$\downarrow$ \\
    \midrule
    0 & 0 & 70.5 & 63.1 & 45.7 & \textbf{2043} & 13048 & 833 \\ % ID-356
    1 & 0 & 71.2 & 72.9 & \textbf{47.2} & 2374 & \textbf{12599} & 627 \\ % ID-357
    1 & 1 & 71.2 & \textbf{73.8} & \textbf{47.2} & 2341 & 12620 & \textbf{587} \\ % ID-358
    % 1 & 5 & 71.4 & 74.3 & 47.2 & 2243 & 12670 & 569 \\ % ID-359
    1 & 10 & \textbf{71.3} & \textbf{73.8} & 46.6 & 2190 & 12742 & 588 \\ % ID-351
    \bottomrule
    \end{tabular}
    }
\end{table}
\noindent \textbf{Long-term association.} 
As shown in Table~\ref{table:long_term}, introducing long-term association into SGT significantly increases IDF1. We use $\mathit{age}_{min}$ of 10 frames so that only the stable tracklets are stored and false positive recovery cases are avoided as much as possible. When the objects are fully occluded as shown in Figure~\ref{fig:sgt_occlusion_longterm}, SGT fails to track and recover them. However, SGT can match them without a motion model when they reappear.

\begin{table}[t!]
    \centering
    \caption{Robustness of $K$. Memory and time taken for training, MOT performance and speed are measured on different $K$ values.}
    \label{table:topk_robustness}
    \resizebox{\linewidth}{!}{
    \begin{tabular}{ @{} c c | c c c | c c @{} }
    \toprule
    $K$ (Train) & $K$ (Test) & MOTA$\uparrow$ & IDF1$\uparrow$ & FPS$\uparrow$ &  Memory (GB) & Time (hr) \\
    \midrule
    100 & 50 &  71.3 & 73.8 & 23.6 & 13.5 & 3 \\ % ID-346
    100 & 100 & 71.3 & 73.8 & 23.5 & 13.5 & 3 \\ % ID-349 
    100 & 300 & 71.5 & 72.8 & 22.2 & 13.5 & 3 \\ % ID-347
    100 & 500 & 71.4 & 72.9 & 21.8 & 13.5 & 3 \\ % ID-348
    300 & 300 & 71.2 & 75.2 & 22.0 & 14.4 & 3.4 \\ % ID-344 & ID-339
    500 & 500 & 71.8 & 73.4 & 21.9 & 15.6 & 3.8 \\ % ID-350 & ID-340
    \bottomrule
    \end{tabular}}
\end{table}
\noindent \textbf{Robustness of $K$ value.}
We validate the robustness of $K$ in SGT by training with different $K$ values (\eg,~100, 300, 500) and inference with unseen $K$ values (\eg,~50, 300, 500) using the model trained with $K=100$. As shown in Table~\ref{table:topk_robustness}, consistent tracking performance is observed across different $K$ values for training and unseen $K$ values. Memory and time consumption for training and inference speed (FPS) are only slightly affected by increasing $K$.

\begin{table}[t!]
    \centering
    \caption{
    % Ablation study of the number of message passing iterations in GNN.
    Effect of the number of GNN iterations.
    }
    \label{table:gnn_iter}
    \resizebox{0.7\linewidth}{!}{
    \begin{tabular}{ @{} c |c c c c c c @{}}
    \toprule
    $N_{iter}$ & MOTA$\uparrow$ & IDF1$\uparrow$ & MT$\uparrow$ & FP$\downarrow$ & FN$\downarrow$ & IDS$\downarrow$\\
    \midrule
    0 & 67.3 & 71.2 & 42.2 & 2538 & 14547 & 578 \\ % ID-308
    1 & 70.9 & 73.2 & 46.0 & 2572 & 12571 & 604 \\ % ID-309
    2 & 71.0 & 73.4 & \textbf{48.4} & 2578 & \textbf{12516} & \textbf{583} \\ % ID-310
    3 & \textbf{71.3} & \textbf{73.8} & 46.6 & \textbf{2190} & 12742 & 588 \\ % ID-311
    % 4 & 70.3 & 72.0 & 47.5 & 2458 & 12966 & 629 \\ % ID-312
    \bottomrule
    \end{tabular}
    }
\end{table}
\noindent \textbf{Number of GNN iterations.}
As shown in Table~\ref{table:gnn_iter}, FN is much higher when GNN is not yet used for updating edge and node features. Also, more GNN iteration improves MOTA and IDF1. This trend proves that the higher-order relational features are more effective in learning spatio-temporal consistency to perform detection recovery than the pairwise relational features ($N_{iter}=0$).

\begin{table}[t!]
\caption{Effect of choosing different relational features for initializing edge feature.}
\label{table:choice_init_features_edge}
\setlength{\tabcolsep}{3pt}
\centering
\resizebox{0.44\textwidth}{!}{
    \begin{tabular}{c | c c c c c}
        \toprule
        Combination of features & MOTA$\uparrow$ & IDF1$\uparrow$ & FP$\downarrow$ & FN$\downarrow$ & IDS$\downarrow$  \\
        \midrule
        x, y, w, h, IoU, Sim & \textbf{71.3} & \textbf{73.8} & 2190 & \textbf{12742} & \textbf{588} \\
        x, y, w, h, IoU & 69.9 & 71.5 & 1953 & 13508 & 821 \\
        x, y, w, h & 70.4 & 70.6 & 2113 & 13205 & 704 \\
        x, y & 69.6 & 70.3 & \textbf{1691} & 13899 & 837 \\
        IoU, Sim & 69.6 & 72.1 & 2066 & 13272 & 640 \\
        \bottomrule
    \end{tabular}
}
\end{table}
\noindent \textbf{Design of edge feature.}
As stated in Eq. 1 of the main paper, we use difference of center coordinates, ratio of width and height, IoU, and cosine similarity to initialize the edge features. Here, both position and appearance relational features are included in the edge features. As shown in Table~\ref{table:choice_init_features_edge}, SGT achieves the best by utilizing all of them.

\section{Conclusion and Future Works}
\label{sec:conclusion}
Partial occlusion in a video leads to low-confident detection output. Existing online MOT models suffer from missed detections since they use only high-scored detections for tracking. This paper presents SGT, a novel online graph tracker that is jointly trained with a detector and recovers the missed detections by tracking top-$K$ scored detections. We also show that pseudo labeling is critical to training SGT and adaptive feature smoothing is a simple but effective inference technique. SGT captures the object-level spatio-temporal consistency in video data by exploiting higher-order relational features of objects and background patches from the current and past frames. SGT outperforms recent online MOT models in MOTA on MOT16/17/20, but particularly shows a large improvement in MOTA on MOT20 which is vulnerable to missed detections due to occlusion caused by severe crowdedness. Our effective detection recovery method contributes to the outstanding performance of SGT as demonstrated in the extensive ablation experiments. Future works will exploit longer temporal cues and model the spatio-temporal relationship of non-human objects (\eg,~vehicles). We hope SGT inspires MOT community to further explore graph tracker and detection recovery by tracking framework for online setting.

\section*{Acknowledgement}
This work was partially supported by a Research Impact Fund project (R6003-21) and a Theme-based Research Scheme project (T41-603/20R) funded by the Hong Kong Government.

{\small
\bibliographystyle{ieee_fullname}
\bibliography{egbib}
}

\clearpage

\appendix

\section*{Appendix}

In this appendix, we provide additional experiment results, visualization results, and detailed analysis of detection recovery which are not included in the main paper.

\section{Additional Implementation Details}
\label{sec:add_impl_details}

We use shorter epochs when CrowdHuman dataset is not included but only MOT or HiEve datasets are used for training. The models are trained for 30 epochs and the learning rate is dropped from $2e^{-4}$ to $2e^{-5}$ at 20 epoch. The same loss weights are adopted, but we use $w_{edge}$ as 1, instead of 0.1. For inference, the same threshold values are used regardless of whether CrowdHuman~\cite{crowdhuman} dataset is used or not. In the ablation experiments, we deploy the detection threshold values of ($\tau_{init}$, $\tau_D$, $\tau_{D_{low}}$) as (0.5, 0.4, 0.2) for FairMOT~\cite{fairmot} and BYTE~\cite{bytetrack} following the official code of ByteTrack\footnote{https://github.com/ifzhang/ByteTrack}.

Our implementation is based on detectron2 framework\footnote{https://github.com/facebookresearch/detectron2}. Regarding training time of the ablation experiments, with two NVIDIA V100 GPUs, 3 hours are spent on training SGT without CrowdHuman dataset, while 2 days are taken if it is included. We observe that the inference speed is affected by the version of NVIDIA driver and the number of CPUs. The reported inference speed is measured on NVIDIA driver version of 460.73.01 with CUDA version of 10.1.

\section{Additional Experiment Results}

\subsection{MOT Detection Challenge Evaluation Results}
\label{subsec:mot_detection_challenge}

For evaluation metrics of MOT17/20 Detection benchmarks, we choose precision, recall, F1, and average precision (AP)~\cite{mscoco_detection_dataset}. As shown in Table~\ref{table:mot_det_result}, SGT achieves the best in every metric on MOT17/20 Detection benchmarks. 
SGT outperforms GSDT~\cite{gsdt} which is also based on CenterNet~\cite{centernet} and GNNs. 
As stated in Section 2.2 of the main paper, GNNs in GSDT aggregate the current and past features to enhance the current image features. 
However, relational features used for association in GSDT are still limited to pairwise features while relational features in SGT are updated by GNNs to become multi-hop features.
Consequently, low-scored detections are not tracked in GSDT.
The high recall supports the effectiveness of detection recovery in SGT. On the other hand, the high precision indicates that detection recovery is achieved without introducing extra false positives.

\begin{table}[t!]
    \centering
    \caption{Evaluation results of the MOT17/20 Detection benchmarks.} 
    \label{table:mot_det_result}
    \resizebox{\linewidth}{!}{
    \begin{tabular}{ @{} c | c | cccccc @{}}
    \toprule
    Benchmark & Method & AP$\uparrow$ & Recall$\uparrow$ & Precision$\uparrow$ & F1$\uparrow$ \\
    \midrule
    \multirow{4}{*}{\begin{tabular}{c} MOT\\17Det\end{tabular}}
        & FRCNN \cite{fasterrcnn}    & 0.72 & 77.3 & 89.8 & 83.1               \\
        & GSDT \cite{gsdt}           & 0.89 & 90.7 & 87.8 & 89.2               \\
        & YTLAB \cite{ytlab}         & 0.89 & 91.3 & 86.2 & 88.7               \\
        & \textbf{SGT (ours)}        & \textbf{0.90} & \textbf{93.1} & \textbf{92.5} & \textbf{92.8}\\ 
    \midrule
    \multirow{3}{*}{\begin{tabular}{c} MOT\\20Det\end{tabular}}
        & ViPeD20 \cite{viped} & 0.80 & 86.5 & 68.1 & 76.2 \\
        & GSDT \cite{gsdt} & 0.81 & 88.6 & 90.6 & 89.6 \\
        & \textbf{SGT (ours)} & \textbf{0.90} & \textbf{91.6} & \textbf{92.6} & \textbf{92.1}  \\ 
    \bottomrule
    \end{tabular}
    }
\end{table}

\subsection{Visualization Comparison}

\begin{figure*}[t!]
    \centering
    \includegraphics[width=\linewidth]{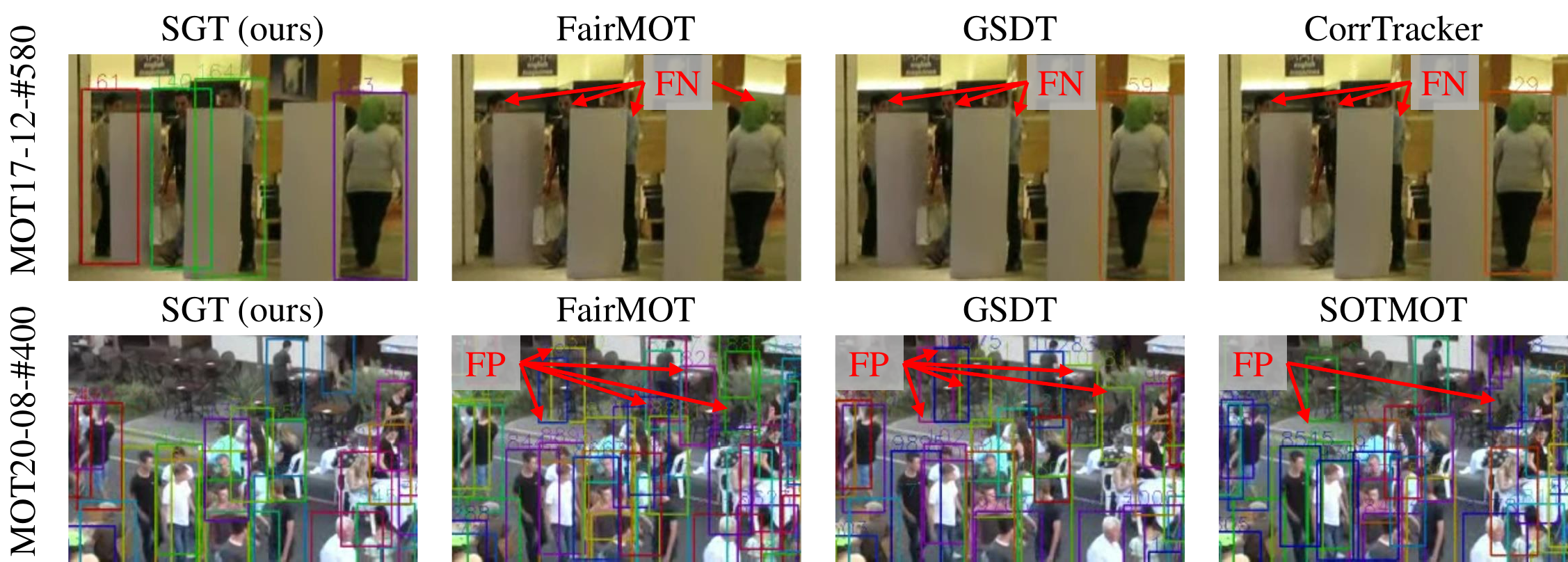}
    \caption{
    Qualitative comparison of CenterNet~\cite{centernet} based online JDT models~\cite{fairmot, gsdt, corrtracker, sotmot} on the MOT17/20 test datasets.
    }
    \label{fig:sgt_qual_comp}
\end{figure*}
Figure~\ref{fig:sgt_qual_comp} shows qualitative comparison between SGT and others~\cite{fairmot, sotmot, gsdt, corrtracker} using the same detector. In MOT17, the partially occluded people have low detection confidence score and they are not used as tracking candidates in FairMOT, GSDT, and CorrTracker since they only use high-scored detections for tracking. As a result, these occluded people are missed.
In MOT20, existing methods use lower detection threshold ($\tau_D$) than MOT17 due to frequent occlusions in MOT20. However, they suffer from FPs while SGT does not have such FPs without missed detections.

\subsection{Additional Ablation Experiments}

\begin{table}[t!]
    \centering
    \caption{Ablation study of threshold and top-$K$ for choosing detections for tracking candidates. CrowdHuman dataset~\cite{crowdhuman} is additionally used for training.}
    \label{table:mot17_thresh_vs_topk}
    \resizebox{\linewidth}{!}{
    \begin{tabular}{ @{} c c | c c c c c c @{}}
        \toprule
        train & inference & MOTA$\uparrow$ & IDF1$\uparrow$ & MT$\uparrow$ & FP$\downarrow$ & FN$\downarrow$ & IDS$\downarrow$  \\
        \midrule
        $K=300$ & $K=300$ & 73.8 & 74.7 & 52.5 & 2620 & 11047 & 476 \\
        $K=300$ & $K=50$ & 73.8 & 74.0 & 52.5 & 2531 & 11159 & 474 \\
        $K=300$ & $\tau_D=0.01$ & \textbf{74.1} & \textbf{74.9} & \textbf{53.4} & 2492 & \textbf{11050} & \textbf{459} \\
        $K=300$ & $\tau_D=0.1$ & 73.5 & 73.4 & 51.6  & 2331 & 11398 & 604 \\
        $K=300$ & $\tau_D=0.3$ & 72.8 & 72.4 & 50.7 & \textbf{2256} & 11605 & 843 \\
        \midrule
        $K=100$ & $K=300$ & 74.1 & \textbf{76.5} & \textbf{53.1} & 2544 & 10971 & 460 \\
        $K=100$ & $K=50$ & \textbf{74.2} & 75.9 & \textbf{53.1} & 2505 & 11000 & 458\\
        $K=100$ & $\tau_D=0.01$ & \textbf{74.2} & 76.2 & \textbf{53.1} & 2530 & \textbf{10961} & \textbf{451} \\
        $K=100$ & $\tau_D=0.1$ & 74.0 & 75.9 & 52.2 & 2391 & 11146 & 538\\
        $K=100$ & $\tau_D=0.3$ & 73.3 & 73.6 & 51.0 & \textbf{2348} & 11313 & 795 \\
        \midrule
        $\tau_D=0.01$ & $K=300$ & \textbf{74.4} & 76.0 & \textbf{53.4} & 2542 & \textbf{10774} & 498 \\
        $\tau_D=0.01$ & $K=50$ & 74.3 & \textbf{77.2} & 53.1 & 2475 & 10920 & \textbf{485} \\
        $\tau_D=0.01$ & $\tau_D=0.01$ & \textbf{74.4} & 76.2 & 53.4 & 2543 & 10780 & 500 \\
        $\tau_D=0.01$ & $\tau_D=0.1$ & 73.2 & 73.1 & 52.2 & 2438 & 11034 & 1002 \\
        $\tau_D=0.01$ & $\tau_D=0.3$ & 71.9 & 71.4 & 51.0 & \textbf{2348} & 11344 & 1480 \\
        \midrule
        $\tau_D=0.1$ & $K=300$ & 73.9 & \textbf{76.0} & 52.5 & 2851 & 10764 & \textbf{470} \\
        $\tau_D=0.1$ & $K=50$ & \textbf{74.0} & \textbf{76.0} & \textbf{53.1} & \textbf{2807} & 10745 & 484 \\
        $\tau_D=0.1$ & $\tau_D=0.01$ & \textbf{74.0} & 75.7 & 52.5 & 2842 & \textbf{10722} & 496 \\
        $\tau_D=0.1$ & $\tau_D=0.1$ & 73.9 & 75.5 & 52.8 & 2849 & 10789 & 486 \\
        $\tau_D=0.1$ & $\tau_D=0.3$ & 73.7 & 74.5 & 52.5 & 2885 & 10804 & 541 \\
        \bottomrule
    \end{tabular}
}
\end{table}
\noindent \textbf{Top-$K$ vs Detection threshold.}
In Table 6 of the main paper, the robustness of top-$K$ sampling in SGT is shown by the consistent performance with a wide range of $K$ values and different $K$ values for training and inference. Here, we experiment about using a low value of detection threshold, $\tau_D$, that is an alternative option for choosing tracking candidates instead of top-$K$ sampling. 

According to Table~\ref{table:mot17_thresh_vs_topk}, when SGT is trained with tracking candidates sampled by top-$K$ whose K is 100 or 300, it shows the consistent performance with both $K=\{50, 300\}$ and $\tau_D=\{0.01, 0.1\}$ while marginal degradation is observed with $\tau_D=0.3$. On the other hand, large drop in MOTA and IDF1 is shown when SGT is trained with $\tau_D=0.01$ and evaluated with $\tau_D=0.3$. In contrast, SGT trained with $\tau_D=0.1$ shows the consistent performance when $\tau_D=0.3$ is used for selecting tracking candidates for inference. 

Based on the results, detection threshold can be viewed as the hyperparameter that should be carefully tuned. Although $K$ is also the hyperparameter, $K$ is more intuitive value representing the maximum number of objects to be tracked and is easier to be decided than $\tau_D$ which is affected by many factors (\eg,~model architecture, training method). For this reason, we adopt top-$K$ sampling method to include low-scored detections in SGT.

\begin{table}[t!]
    \centering
    \caption{Performance comparison of FairMOT~\cite{fairmot} and SGT with different backbone networks: ResNet-18/101~\cite{resnet}, DLA-34~\cite{dla}, and hourglass-104~\cite{hourglass}. The models are trained using the extra CrowdHuman~\cite{crowdhuman} dataset.}
    \label{table:backbone_ablation}
    \resizebox{\linewidth}{!}{
    \begin{tabular}{ @{} c c | c c c c c c c c @{}}
        \toprule
        Model & Backbone & MOTA$\uparrow$ & IDF1$\uparrow$ & MT$\uparrow$ & ML$\downarrow$ & FP$\downarrow$ & FN$\downarrow$ & IDS$\downarrow$\\
        \midrule
        FairMOT \cite{fairmot} & Res-18 & 66.1 & \textbf{69.9} & 40.1 & 20.4 & \textbf{2036} & 16029 & \textbf{265} \\ % ID-213
        SGT (ours) & Res-18 & \textbf{68.4} & 69.3 & \textbf{47.2} & \textbf{15.6} & 2359 & \textbf{14086} & 659 \\ % ID-29
        \midrule
        FairMOT \cite{fairmot} & Res-101 & 70.2 & 72.2 & 47.8 & 14.5 & \textbf{2545} & 13178 & \textbf{364} \\ % ID-214
        SGT (ours) & Res-101 & \textbf{71.1} & \textbf{72.4} & \textbf{53.4} & \textbf{12.4} & 3197 & \textbf{11678} & 720 \\ % ID-28
        \midrule
        FairMOT \cite{fairmot} & DLA-34 & 72.2 & 74.7 & 47.8 & 18.0 & 2660 & 12025 & \textbf{336} \\ % ID-215
        SGT (ours) & DLA-34 & \textbf{74.2} & \textbf{76.3} & \textbf{53.1} & \textbf{13.3} & \textbf{2514} & \textbf{10978} & 451 \\ % ID-328
        \midrule
        FairMOT \cite{fairmot} & HG-104 & 74.4 & \textbf{77.1} & 54.0 & 12.1 & \textbf{2636} & 10844 & \textbf{344} \\ % ID-9
        SGT (ours) & HG-104 & \textbf{74.8} & \textbf{77.1} & \textbf{56.0} & \textbf{10.9} & 2713 & \textbf{10454} & 428 \\ % ID-64
        \bottomrule
    \end{tabular}}
\end{table}
\noindent \textbf{Different backbone networks.}
Table~\ref{table:backbone_ablation} shows the performance comparison of SGT and FairMOT~\cite{fairmot} with different backbone networks~\cite{resnet, dla, hourglass} used in CenterNet~\cite{centernet}. For hourglass backbone, we use the image size of $(H, W)=(640, 1152)$, instead of $(608, 1088)$ since only a multiple of 128 is allowed. SGT achieves lower FN and ML, and higher MT and MOTA than FairMOT across all backbone networks. In other words, SGT has less missed detections and more long-lasting tracklets than FairMOT. This result is corresponding to our motivation of detection recovery by tracking in SGT. Especially, SGT shows larger improvement in MOTA with the small backbone networks (\eg,~resnet18 and dla34). When MOT models are deployed with the limited resource of hardware, SGT can be served as an effective solution.

\begin{table}[t!]
\centering
\caption{Effect of the number of the edges for each criterion. Center distance, IoU, and cosine similarity are three criteria used in SGT.}
\label{table:num_edges}
\resizebox{\linewidth}{!}{
    \begin{tabular}{ @{} c | c c c c c c @{}}
        \toprule
        \#edges per criterion & MOTA$\uparrow$ & IDF1$\uparrow$ & MT$\uparrow$ & FP$\downarrow$ & FN$\downarrow$ & IDS$\downarrow$  \\
        \midrule
        5 & 71.0 & 72.8 & 47.2 & 2450 & 12590 & 624 \\ % ID-407
        10 & \textbf{71.3} & \textbf{73.8} & 46.6 & \textbf{2190} & 12742 & \textbf{588} \\ % ID-414
        20 & 71.1 & 73.6 & \textbf{47.8} & 2308 & \textbf{12535} & 755 \\ % ID-412
        \bottomrule
    \end{tabular}
}
\end{table}
\noindent \textbf{The number of edges.}
In SGT, nodes across frames are sparsely connected if only they are close in either Euclidean or feature space. Specifically, $n^i_{t1}$ is connected to the nodes of $N_{t2}$ using three criteria: center distance, IoU, and cosine similarity. We choose nodes for each criterion and remove the duplicates. Table~\ref{table:num_edges} shows the result of experimenting with different number of nodes for each criterion. Although the best performance is achieved with 10, there is only marginal decrease in MOTA and IDF1 with 5 and 20. Thus, this is also robust hyperparameter.

\begin{figure}
    \centering
    \includegraphics[width=1.0\linewidth]{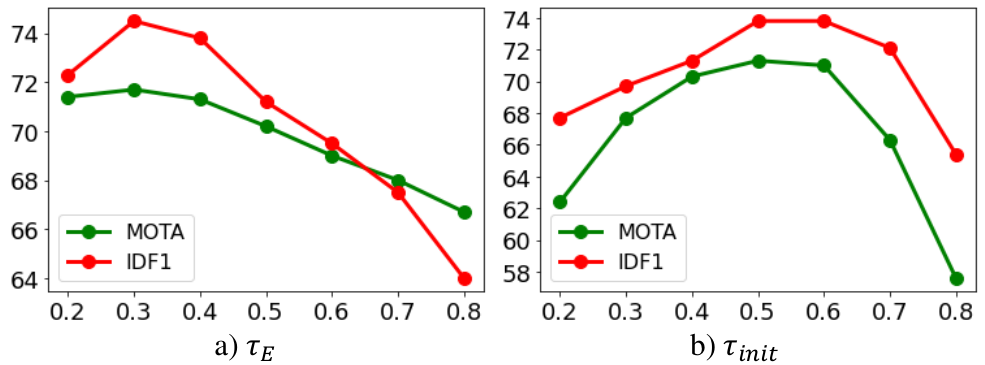}
    \caption{Sensitivity analysis of $\tau_E$ and $\tau_{init}$.}
    \label{fig:sgt_thresh_tuning}
\end{figure}
\noindent \textbf{Sensitivity of $\tau_{init}$ and $\tau_E$.} The robustness of $K$, which is the number of tracking candidates, has been shown through the extensive ablation experiments. Here, we measure the sensitivity of $\tau_{init}$ and $\tau_E$ as well. As shown by Figure~\ref{fig:sgt_thresh_tuning}, $\tau_{init}$ is a sensitive threshold value since it decides initialization of new tracklets. On the other hand, $\tau_E$ is the minimum edge score used for matching. It is robust within the range between 0.2 and 0.4 since correct matching may have high edge score and the node classifier prevents false positive matching.

\section{Analysis of Detection Recovery}
\label{sec:analysis_detection_recovery}

\begin{table}[t!]
\caption{Ratio of recovered detections over all detections in each sequence of MOT17/20 test datasets.}
\label{table:ratio_recovery}
\centering
\resizebox{0.7\linewidth}{!}{
\begin{tabular}{@{} c | c|c @{}}
\toprule
Benchmark & Sequence     & Recovery Ratio (\%) \\
\midrule
\multirow{7}{*}{\begin{tabular}{c} MOT17 \end{tabular}}
& MOT17-01          & 12.0                                               \\
& MOT17-03          & 1.8                                               \\
& MOT17-06          & 6.8                                               \\
& MOT17-07          & 10.0                                              \\
& \textbf{MOT17-08} & \textbf{14.6}                                     \\
& MOT17-12          & 10.3                                              \\
& MOT17-14          & 12.6                                              \\
\midrule
\multirow{4}{*}{\begin{tabular}{c} MOT20 \end{tabular}}                 
& MOT20-04          & 3.8                                               \\
& MOT20-06          & 29.0                                              \\
& MOT20-07          & 4.9                                              \\
& \textbf{MOT20-08} & \textbf{35.4}                                     \\
\bottomrule
\end{tabular}
}
\end{table}
\begin{table}[t!]
\centering
\caption{Evaluation result per sequence of MOT17/20 test dataset. We choose one with high recovery ratio and the other one with low recovery ratio.}
\label{table:seq_results}
\resizebox{\linewidth}{!}{
    \begin{tabular}{@{} l |ccccccc @{}}
    \toprule
    Method & MOTA$\uparrow$ & IDF1$\uparrow$ & MT$\uparrow$ & ML$\downarrow$ & FP$\downarrow$ & FN$\downarrow$ & IDS$\downarrow$ \\
    \midrule \midrule
    \multicolumn{8}{c}{MOT17-06 (6.8\% recovery)} \\
    \midrule
    SGT (Ours) & 65.5 & 63.2 & \textbf{48.2} & \textbf{12.2} & 942 & \textbf{2917} & 210 \\
    FairMOT \cite{fairmot} & 64.1 & 65.9 & 40.1 & 18.5 & 526 & 3533 & 176 \\
    GSDT \cite{gsdt} & 63.0 & 62.0 & 40.1 & 21.2 & 681 & 3500 & 180 \\
    CorrTracker \cite{corrtracker} & \textbf{66.2} & \textbf{68.2} & 41.0 & 17.1 & \textbf{465} & 3346 & \textbf{171} \\
    \midrule
    \multicolumn{8}{c}{MOT17-08 (14.6\% recovery) }  \\
    \midrule
    SGT (Ours) & \textbf{52.6} & 44.0 & \textbf{32.9} & \textbf{14.5} & 1076 & \textbf{8546} & 347 \\
    FairMOT \cite{fairmot} & 42.2 & 42.0 & 22.4 & 28.9 & \textbf{776} & 11191 & \textbf{237} \\
    GSDT \cite{gsdt} & 44.0 & 40.5 & 26.3 & 22.4 & 991 & 10523 & 323 \\
    CorrTracker \cite{corrtracker} & 49.9 & \textbf{46.7} & 25.0 & 17.1 & 1137 & 9201 & 250 \\
    \midrule
    \multicolumn{8}{c}{MOT20-07 (4.9\% recovery)} \\
    \midrule
    SGT (Ours) & \textbf{77.9} & \textbf{71.4} & \textbf{76.6} & 2.7 & 2277 & 4774 & \textbf{254} \\
    FairMOT \cite{fairmot} & 75.6 & 70.0 & \textbf{76.6} & \textbf{0.9} & 2988 & \textbf{4770} & 333 \\
    GSDT \cite{gsdt} & 75.0 & 68.1 & 64.0 & 1.8 & \textbf{1870} & 6115 & 282 \\
    SOTMOT \cite{sotmot} & 72.6 & 71.2 & \textbf{76.6} & 2.7 & 3675 & 5066 & 317 \\
    \midrule
    \multicolumn{8}{c}{MOT20-08 (35.4\% recovery)} \\
    \midrule
    SGT (Ours) & \textbf{54.1} & 54.5 & 26.7 & 26.2 & \textbf{2434} & 32625 & \textbf{468} \\
    FairMOT \cite{fairmot} & 27.0 & 49.5 & \textbf{41.9} & \textbf{14.7} & 32104 & \textbf{23447} & 981 \\
    GSDT \cite{gsdt} & 39.4 & 48.9 & 22.5 & 32.5 & 9916 & 36420 & 608 \\
    SOTMOT \cite{sotmot} & 43.1 & \textbf{55.1} & 35.6 & 19.9 & 16025 & 27216 & 863 \\
    \bottomrule
    \end{tabular}
}
\end{table}

\subsection{Ratio of Recovered Detections}
According to Table~\ref{table:ratio_recovery}, MOT17-08 and MOT20-08 are the sequences that SGT outputs the highest ratio of recovered detections whose confidence score is lower than $\tau_{init}=0.5$. 
Table~\ref{table:seq_results} shows that SGT surpasses CorrTracker in MOTA by 2.7\% in MOT17-08 while SGT shows lower MOTA than CorrTracker in MOT17-06 whose recovery ratio is low. In MOT20, similar trend is observed that SGT achieves larger improvement of MOTA in MOT20-08 than MOT20-07.

\begin{figure*}[t!]
    \centering
    \includegraphics[width=1.0\textwidth]{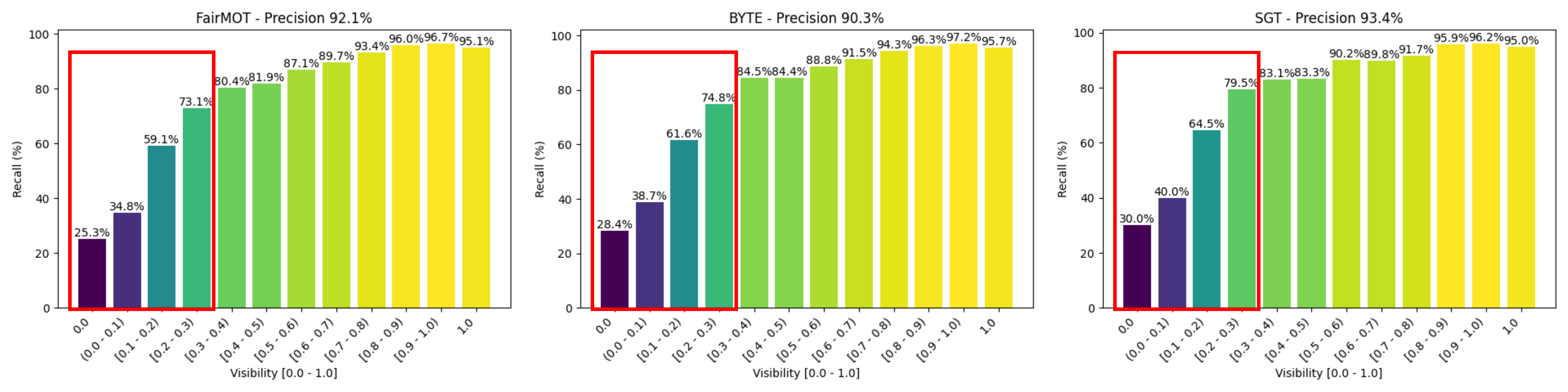}
    \caption{
    Recall ratio comparison of FairMOT~\cite{fairmot}, BYTE~\cite{bytetrack} on top of FairMOT~\cite{fairmot}, and SGT for different visibility levels of objects in MOT17 validation dataset.}
    \label{fig:sgt_recall_per_visibility}
\end{figure*}

\begin{figure*}[t!] 
    \centering
    \includegraphics[width=1.0\textwidth]{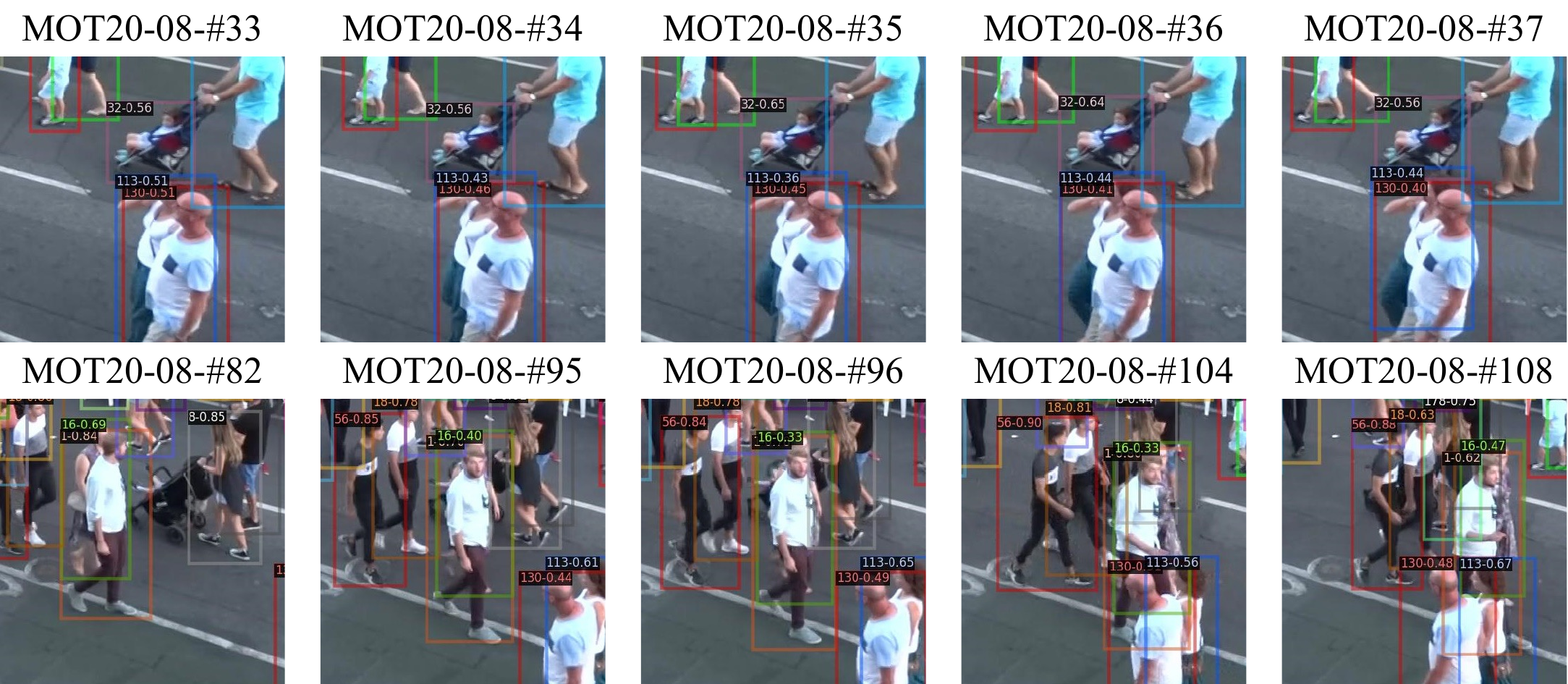}
    \caption{Detection recovery cases in MOT20 test dataset~\cite{mot20}. We show the annotation of each bounding box in the format of ``\{id\}-\{detection score\}''. The tracklets in the red circles are recovered in the next frames.
    }
    \label{fig:sgt_mot20-recovery}
\end{figure*}

\subsection{Recall per Visibility Level}
We measure the recall ratio for different visibility levels of objects and compare them of different models as shown in Figure~\ref{fig:sgt_recall_per_visibility}. Both BYTE~\cite{bytetrack} and SGT show higher recall value for low visibility levels than FairMOT~\cite{fairmot} since they perform association of low-scored detections. When objects are almost invisible with the visibility in the range of (0.0, 0.3), SGT outperforms BYTE in terms of the recall. These results indicate that SGT successfully tracks the low-scored detections caused by occlusion, and SGT is robust against partial occlusion. Also, the effectiveness of node classifier preventing false positive recovery is demonstrated through higher precision value of SGT than that of BYTE.

\subsection{Visualization of Recovered Detections}
Figure~\ref{fig:sgt_mot20-recovery} shows the cases of detection recovery in the MOT20 test dataset. In the first row, people indicated by the blue and brown bounding boxes occlude each other. In the frame \#33, their detection scores are higher than $\tau_{init}$ which is detection threshold value to initialize new tracklet. However, from frame \#34 to \#37, their detection scores are between 0.3 or 0.4 which are lower than $\tau_{init}$, nevertheless, SGT successfully tracks them. If only high-scored detections are used for association, they are failed to track, leading to missed detections and disconnected tracklets. The second row of Figure~\ref{fig:sgt_mot20-recovery} is another example of detection recovery.

% \begin{figure*}
%     \centering
%     \includegraphics[width=1.0\linewidth]{pics/sgt/ids_case.pdf}
%     \caption{Example of ID switch caused by non-human occluder.}
%     \label{fig:id_switch_mot17}
% \end{figure*}

\begin{figure}
    \centering
    \includegraphics[width=1.0\linewidth]{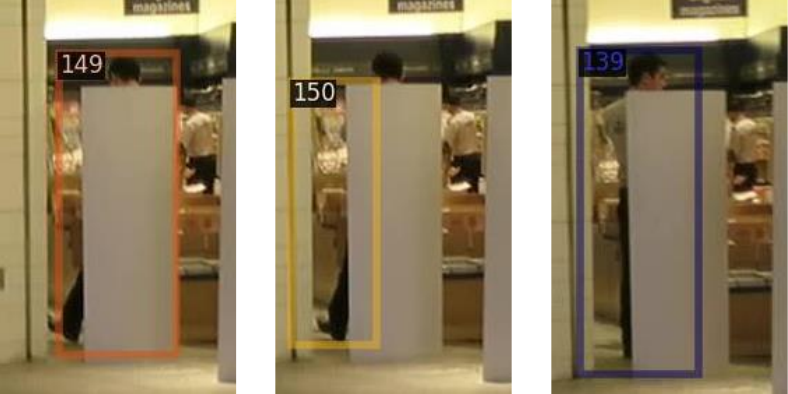}
    \caption{Example of ID switch caused by non-human occluder.}
    \label{fig:id_switch_mot17}
\end{figure}

\section{Discussion}
\noindent \textbf{High IDS in MOT16/17.}
In Section 4.2 of the main paper, we stated that non-human occluders in MOT16/17 result in high IDS in MOT16/17 compared to low IDS in MOT20. Figure~\ref{fig:id_switch_mot17} shows the example whose video is taken from a department store, where non-human occluders commonly exist.

% % \clearpage
% \newpage

\end{document}